\PassOptionsToPackage{dvipsnames}{xcolor}
\documentclass{article} %
\usepackage{iclr2027_conference,times}

\usepackage{amsmath,amsfonts,bm}

\def\eqref#1{equation~\ref{#1}}

\def\1{\bm{1}}

\DeclareMathAlphabet{\mathsfit}{\encodingdefault}{\sfdefault}{m}{sl}
\SetMathAlphabet{\mathsfit}{bold}{\encodingdefault}{\sfdefault}{bx}{n}

\usepackage[normalem]{ulem}   %
\usepackage{xcolor}
\usepackage{amssymb}
\usepackage{amsthm} 
\usepackage{multirow}
\usepackage{xspace}
\usepackage{booktabs,fontawesome5,pifont,xcolor,colortbl}
\definecolor{reqbg}{RGB}{253,230,224}
\definecolor{freebg}{RGB}{226,245,230}

\definecolor{video}{RGB}{0, 114, 178}
\definecolor{control}{RGB}{213, 94, 0}

\usepackage[dvipsnames]{xcolor}

\newcommand{\wmpi}{\mbox{WM\,+\,policy}\xspace}
\newcommand{\vpva}{\mbox{VP\,+\,V2A}\xspace}

\definecolor{propold}{gray}{0.55}      %
\definecolor{propnew}{HTML}{B4544A}    %

\usepackage{tcolorbox}
\usepackage{changepage}

\usepackage{tabularx}
\usepackage{tikz}
\usetikzlibrary{decorations.pathreplacing}
\usepackage{listings}
\lstdefinestyle{appcode}{
  basicstyle=\ttfamily\scriptsize,
  breaklines=true,
  breakindent=0pt,
  columns=fullflexible,
  keepspaces=true,
  showstringspaces=false,
  frame=single,
  rulecolor=\color{black!25},
  backgroundcolor=\color{black!2},
  framesep=3pt,
  xleftmargin=3pt,
  xrightmargin=3pt,
  aboveskip=4pt,
  belowskip=4pt,
  literate={—}{{---}}1 {’}{{'}}1
}

\newtcolorbox{episodebox}{
  colback=black!2,
  colframe=black!25,
  colbacktitle=black!7,
  coltitle=black,
  title={Data collection episode},
  fonttitle=\bfseries,
  boxrule=0.6pt,
  arc=2pt,
  left=4pt,
  right=4pt,
  top=3pt,
  bottom=3pt,
  before skip=4pt,
  after skip=4pt
}

\newtcolorbox{phasebox}[2]{
  colback=#1!4,
  colframe=#1!45,
  colbacktitle=#1!10,
  coltitle=#1,
  title={#2},
  fonttitle=\bfseries,
  boxrule=0.5pt,
  arc=2pt,
  left=3pt,
  right=3pt,
  top=2pt,
  bottom=2pt,
  before skip=3pt,
  after skip=3pt
}

\newcommand{\methodname}{\text{Mirror}}
\newcommand{\website}{\href{https://wilcoxsonm21.github.io/mirror/}}

\usepackage{graphicx}
\usepackage[pagebackref=true,breaklinks=true,letterpaper=true,colorlinks,bookmarks=false,citecolor=BlueViolet,linkcolor=Bittersweet]{hyperref}
\usepackage{url}
\usepackage{algorithm}
\usepackage{algpseudocode}
 \usepackage{enumitem}
\usepackage{amsmath}
\usepackage{booktabs}
\usepackage{capt-of}
\usepackage{wrapfig}

\title{Self-Improving Video-to-Action Models}

\author{Hyunwoo Ryu\thanks{Equal contribution.}\qquad Eric M. Chen$^*$\qquad Max Wilcoxson$^*$\qquad Vincent Sitzmann \\
Computer Science \& Artificial Intelligence Laboratory\\
Massachusetts Institute of Technology\\
Cambridge, MA 02139, USA \\
\texttt{\{hwryu,echen01,mwil,sitzmann\}@mit.edu} \\
}

\iclrfinalcopy %
\begin{document}

\maketitle

\begin{abstract}
Video generative models are an emerging tool for embodied agents: they can serve as scalable visual planners, generating videos that depict how to complete previously unseen tasks.
However, executing these plans typically requires a video-to-action or ``inverse dynamics'' model trained on large, human-collected video-action datasets, severely limiting scalability.
We introduce \methodname{}, a self-supervised approach that learns video-to-action mappings through environment interactions, without prior action data or hand-crafted rewards.
\methodname{} learns by attempting to follow a visual plan by executing its best guess of the underlying actions in the environment. \methodname{} then trains on the resulting video-action pairs, incrementally self-improving and following the visual planner more closely.
Across Atari, OGBench, and DeepMind Control, \methodname{} outperforms leading learning-from-observation methods with learned video plans, and achieves near-expert performance with oracle plans. On Minecraft BASALT, it outperforms methods with access to expert actions or hand-engineered task knowledge. In Minecraft survival, it is the first method to acquire an iron pickaxe through self-improvement alone, without rewards or expert action labels. In the offline setting, with expert action labels but no environment interaction, it sets a new state of the art in obtaining diamonds.
Our results suggest that advances in video models can translate into increasingly capable agents without a corresponding need for hand-crafted rewards or large collections of expert action data.
Please find video and other results on our \website{website.}
\vspace{-1\baselineskip}

\end{abstract}

\section{Introduction}
Video models are emerging as scalable learners of the physical world. Trained on vast collections of unstructured web video, they absorb knowledge about objects, environments, human behavior, and how tasks unfold over time. 
Recent work has begun to harness this knowledge for robot control by using video models as \emph{visual planners} (e.g., ~\citealp{du2023unipi,chen2025lvp,ye2026world,rhoda2026dva}). Given past frames and optional task conditioning, these models generate a video depicting how the task could be completed. An embodiment-specific video-to-action (V2A) or ``inverse dynamics'' model then translates the generated visual trajectory into robot actions. This has yielded impressive generalization across unseen tasks, objects, environments, and embodiments.

However, training the V2A model still requires significant amounts of expert-collected, paired video-action data from oracle motion planners or human teleoperation. 
Moreover, agents cannot easily generate such video-action data by rolling out random policies in the environment, as critical environment transitions---such as executing a robot grasp or crafting a diamond pickaxe in Minecraft---are unreachable by a random policy. 
Today, video planners thus inherit the same scalability bottleneck as video-language-action models: a reliance on expensive expert teleoperation data.

In this paper, we introduce \methodname{}, a \emph{fully self-supervised} method for learning to execute the policy encoded by a video model only by interacting with the world, without any expert video-action data. We assume access to an expert video planner that, given the current agent observation, can generate an expert-level visual plan. Under this assumption, we show that no further expert action supervision or hand-crafted rewards are  necessary to achieve expert-level performance. \methodname{} thus represents a bet on the continued scaling of video models: as visual planners improve, their capabilities can be translated into actions without a corresponding increase in expert action data.

Our key idea is to let a randomly initialized video-to-action model self-improve against the expert video planner. Given the current agent state, we query the planner to generate a video plan. 
The current video-to-action model then predicts the corresponding actions and the agent executes them in the environment. 
Interactions produce new paired video-action data, which is used to improve the video-to-action model. Critically, this data is not random but concentrated at the frontier of the agent's current capability to follow the video plan. 
Repeating this process allows the agent to follow progressively more of the plan until it faithfully reproduces the planner’s behavior. \methodname{} does not require rewards, privileged state information, verifiers, or trajectory filtering. Furthermore, we show that our self-improving approach can outperform training on expert actions.

We evaluate \methodname{} in two settings: (i) the Minecraft survival challenge~\citep{guss2019minerl} and Minecraft BASALT competition tasks~\citep{Shah2021BASALT}; and (ii) standard decision-making benchmarks, including Atari~\citep{bellemare2013atari}, OGBench~\citep{park2025ogbench}, and DeepMind Control~\citep{tassa2018deepmindcontrolsuite}. To the best of our knowledge, no prior method without either hand-crafted rewards or expert action labels achieves non-zero success on these Minecraft tasks. 
\methodname{} outperforms baselines, including those with privileged information, and further exploits the emergent in-context learning of large video models, improving faster than a baseline that cannot leverage context across attempts.

\textbf{Our contributions are summarized as follows:}
\begin{itemize}[leftmargin=*,labelindent=0pt,labelsep=0.4em,nosep,topsep=2pt]
    \item \methodname{}, a scalable agent that learns long-horizon tasks by self-improving against a pre-trained video planner, without task rewards or expert video-action data.
   \item A heavily adapted 14B-parameter NVIDIA Cosmos3 video planner with efficient long-context planning via causal KV-caching. Code and weights will be available soon on our~\website{website}.
    \item Mirror is the first method to achieve non-zero success rates on the BASALT Minecraft suite without expert video-action data or hard-coded actions.
    \item In Minecraft survival, \methodname{} is the first method to obtain an iron pickaxe through self-improvement alone, using neither rewards nor expert action labels. Trained offline with action labels, it sets a new state of the art, obtaining diamonds nearly eight times as often as Dreamer~V4.
    \item On a suite of 24 common RL tasks, \methodname{} outperforms baselines using a generative video planner, and, when paired with an oracle video planner, nearly matches expert returns.
\end{itemize}

\section{Related Work}

\textbf{Video planners.}
A nascent line of work leverages video generative models to generate a visual plan of how a given task should be solved, and then translates it into actions~\citep{du2023unipi,ko2024avdc,liang2024dreamitate,Luo2024GroundingVM,bharadhwaj2024gen2act}.
As video models can be pre-trained on web-scale data, robots can inherit impressive generalization capabilities~\citep{chen2025lvp,li2026vera}.
However, translating videos into robot actions generally requires training on paired video-action data~\citep{du2023unipi,tian2024pidm,pai2025mimicvideo}.
Our agent learns to act from video and self-play in an interactive environment alone, reaching imagined futures without a single action label.

\textbf{Learning from demonstrations.} We focus on imitation from observation, where an agent learns from expert observations without action labels or task rewards~\citep{torabi2018behavioralcloningobservation}. Adversarial approaches match expert observation distributions through learned discriminator rewards~\citep{torabi2019generativeadversarialimitationobservation, liu2023visualimitationlearningpatch, giammarino2024adversarialimitationlearningvisual}, but can be unstable and sensitive to hyperparameters~\citep{huang2024diffusionimitationobservation}. Iterative behavioral cloning alternates between training a V2A model on the agent's own experience and imitating V2A-labeled demonstrations~\citep{torabi2018behavioralcloningobservation}, but collecting informative transitions remains an exploration challenge on long-horizon tasks~\citep{schmidt2024learningactactions}. 
ASH~\citep{schneider2026ash} applied a similar recipe to long-horizon games such as Pokémon and Zelda, training an inverse dynamics model on the agent's own trajectories to pseudo-label internet gameplay videos. 
Latent-action approaches such as LAPO and LAPA still ground their latent actions with action labels or rewards~\citep{schmidt2024learningactactions, ye2025latentactionpretrainingvideos}. \methodname{} instead uses a video planner to guide exploration, collecting V2A training data without action labels, rewards, or adversarial optimization.

\textbf{Minecraft agents.}
Minecraft has become a standard benchmark for long-horizon decision making, ranging from the fuzzy objectives of BASALT~\citep{Shah2021BASALT} to the canonical MineRL challenges~\citep{guss2019minerl}.
Existing agents rely on some combination of expert action demonstrations and hand-designed rewards.
For example, VPT~\citep{VPT2022OpenAI} trains a behavioral cloning (BC) policy using about 2,000 hours of action-labeled gameplay recordings from paid contractors, then fine-tunes the policy with about 230,000 hours of online RL. While Dreamer~V3~\citep{hafner2023dreamerv3} trains a Minecraft RL agent with only 1,400 hours of online RL, they simplify parts of the Minecraft action space and still require hand-designed reward. Dreamer~V4~\citep{hafner2025dreamerv4} likewise uses reward, and also requires expert actions to train a video world model. Pan-4B~\citep{pantograph2026pan1} trains a goal-conditioned RL policy on 500k hours of internet video, yet still post-trains on 2,000 hours of expert action labels. 
To make long-horizon tasks more tractable, many works improve upon Minecraft policies with language models. MineDojo~\citep{fan2022minedojo} trains PPO agents on shaped rewards and a learned video-language critic, while STEVE-1~\citep{lifshitz2023steve1}, DEPS~\citep{wang2023deps}, and JARVIS-1~\citep{wang2023jarvis1} use language-based planning. 

Prior agents that solve long-horizon tasks thus depend on hand-crafted rewards, expert video-action data, or both; \methodname{} instead only assumes access to an expert-level video planner.

\section{Method}
\label{sec:method}
\begin{figure}[t]
    \centering
    \includegraphics[width=1.00\linewidth]{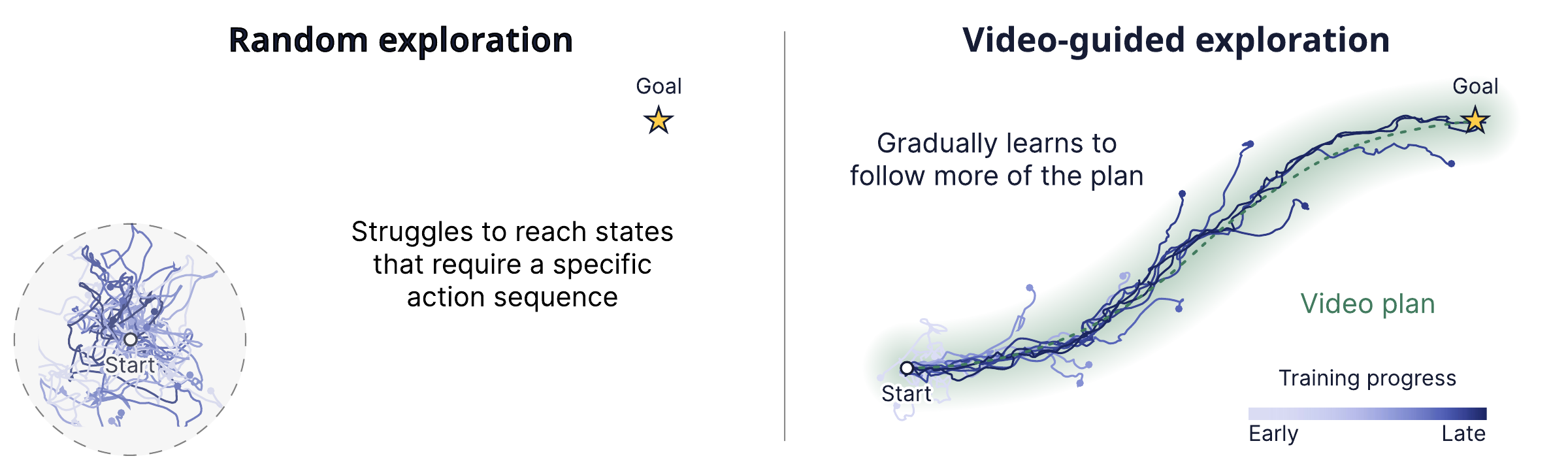}
    \vspace{-1\baselineskip}
    \caption{%
        \textbf{Enabling long-horizon exploration with a video planner.}
        \textbf{Left:} Random exploration struggles to reach later stages of long-horizon tasks that require specific sequences of actions. 
        \textbf{Right:} The video planner guides exploration. As the V2A model learns from achieved transitions, the agent follows the plan more accurately and reaches progressively later stages of the task. These new transitions further improve the model, eventually enabling the agent to execute the full plan. 
    }
    \label{fig:exploration_intuition}
    \vspace{-1\baselineskip}
\end{figure}

We seek a task-conditioned policy $\pi(a_{t}|o_{\leq t}, l_{\mathtt{task}})$, which predicts actions $a_t$ based on past observations $o_{\leq t}$ and a task $l_{\mathtt{task}}$.
We model $\pi$ indirectly with two components: a video planner (VP) and a video-to-action (V2A) model\footnote{V2A models are also commonly referred to as \emph{inverse dynamics models (IDMs)}.}. The VP predicts future observations $\hat{o}_{t+1}$ based on historical observations $o_{\leq t}$ and a task $l_{\mathtt{task}}$, while the V2A model predicts actions $a_t$ given the combined video sequence $(o_{\leq t}, \hat{o}_{t+1})$. Sequentially sampling from the two models yields the agent's policy:
\begin{align*}
a_t \sim  
      p_{\mathtt{V2A}}(\cdot | o_{\leq t}, \hat{o}_{t+1}) \quad \xleftarrow[]{\hspace{1.5em}} \quad  \hat{o}_{t+1} \sim p_{\mathtt{VP}}
    (\cdot\mid o_{\leq t},l_{\mathtt{task}})
\end{align*}
Crucially, every plan is conditioned on the frames the agent \emph{actually} observed, never on its previous plans. The VP therefore always plans from wherever the agent ended up, so deviations from earlier plans can be corrected. For the same reason, errors do not compound over time, which removes the need for anti-drift techniques against exposure bias in autoregressive video generation~\citep{chen2024diffusion,huang2025selfforcing,zhu2026causalforcing}.

This factorization differs from the world-model (WM) and policy decomposition used by VLAs and world-model agents~\citep{hafner2025dreamerv4}, although both model the same joint distribution:
\begin{equation}
\label{eqn:decompose}
\begin{split}
    p(o_{t+1}, a_t \mid o_{\leq t}, l_{\mathtt{task}})
&= \underbrace{p_{\mathtt{VP}}(o_{t+1}\mid o_{\leq t}, l_{\mathtt{task}})}_{\text{action-free}}
  \times \underbrace{p_{\mathtt{V2A}}(a_t \mid o_{\leq t}, o_{t+1})}_{\text{task-agnostic}}
\\
&= \underbrace{p_{\mathtt{WM}}(o_{t+1}\mid o_{\leq t}, a_t)}_{\text{action-bound}}
\times \underbrace{\pi(a_t\mid o_{\leq t}, l_{\mathtt{task}})}_{\text{task-specific}}.
\end{split}
\end{equation}
The two decompositions differ in where the difficult reasoning is learned. Long-horizon tasks require both procedural knowledge of what to do and physical knowledge of how the world responds. In the \wmpi decomposition, the policy must acquire the procedural knowledge and the WM the action-conditioned physical knowledge, both from scarce, embodiment-specific action data. In the \vpva decomposition, the VP learns both from abundant action-free video, which enables web-scale pretraining. This leaves the V2A model a much easier problem: a task-agnostic mapping from observed transitions to actions. It therefore needs far less embodiment-specific data, such as action labels. The \vpva decomposition is thus considerably more scalable, as our experiments (Section~\ref{sec:oracle_survival}) and recent results in robot control~\citep{li2026vera} show.

Our method, \methodname{}, reveals a far more striking consequence of this decomposition: given a capable VP and the ability to interact with the environment, the V2A model needs no action labels at all. Because it is task-agnostic, it can learn from any transition the agent itself produces, while the VP steers the agent toward the transitions the task requires. Section~\ref{sec:V2A} shows how the V2A model can be trained from scratch on the data the agent collects while following these plans, and Section~\ref{sec:VP} describes how we make a generative video model VP practical for an online agent.

\begin{wrapfigure}{r}{0.34\linewidth}
    \vspace{-2.9em}
    \centering
    \includegraphics[width=\linewidth]{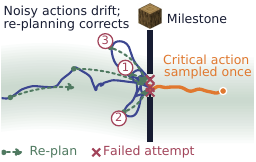}
    \vspace{-1.8em}
    \caption{Re-planning holds the agent at the milestone until noise supplies the critical action.}
    \label{fig:single_attempt}
\end{wrapfigure}
\subsection{Self-improving Video-to-Action Models}
\label{sec:V2A}
Our key insight is to use a pretrained video planner to guide data collection for training the V2A model. The model learns from the agent's own rollouts, without expert action labels. As summarized in Algorithm~\ref{alg:main}, training alternates between data collection and model updates. During data collection, we sample an action from the agent's two-stage policy $\pi$ and add exploration noise $n_t {\sim} \pi_{\mathtt{explore}}$. We execute the resulting action $a_t$ in the environment and store the transition $(o_{\leq t}, a_t, o_{t+1})$ in a replay buffer. To update the model, we sample batches from the buffer and train the V2A model to predict $a_t$ from $(o_{\leq t}, o_{t+1})$. Because the targets are conditioned on the \emph{achieved} observation $o_{t+1}$ rather than the planned one $\hat{o}_{t+1}$, every transition is a valid training example, even when execution deviates from the plan. Further details can be found in Algorithm~\ref{alg:main}.

\begin{algorithm}[t!]
\caption{Learning without action labels: VP-guided self-training of the V2A model}
\label{alg:main}

\begin{algorithmic}
\Require video planner $p_{\mathtt{VP}}$, environment $\mathcal{E}$,
exploration distribution $\pi_{\mathtt{explore}}$,
 horizon $T$.
\State Initialize V2A model $p_{\mathtt{V2A}}$ with parameters $\theta$ from scratch;
replay buffer $\mathcal{D}\gets\emptyset$.
\Statex \hspace{-10pt}\textbf{Repeat until convergence:}
\end{algorithmic}

\begin{adjustwidth}{1em}{0pt}

\begin{phasebox}{video}{Data collection episode}

\begin{algorithmic}
\State Reset $\mathcal{E}$; observe $o_0$;
initialize $\tau\gets(o_0)$ and $t\gets0$
\end{algorithmic}

\begin{adjustwidth}{1em}{0pt}

\begin{algorithmic}
\While{$t < T$}
    \State $\hat{o}_{t+1}
    \sim p_{\mathtt{VP}}
    (\cdot\mid o_{\leq t},l_{\mathtt{task}})$

    \State $a_t
    \sim p_{\mathtt{V2A}}
    (\cdot\mid o_{\leq t},\hat{o}_{t+1};\theta)$ and $n_t \sim \pi_\mathtt{explore}$
    \State $a_t \leftarrow a_t + n_t$

    \State Execute $a_t$ in $\mathcal{E}$ and observe $o_{t+1}$

    \State $\tau \gets \tau :: (a_t,o_{t+1})$ and $t\gets t+1$
\EndWhile
\end{algorithmic}

\end{adjustwidth}

\begin{algorithmic}
\State $\mathcal{D}\gets\mathcal{D}\cup\{\tau\}$
\end{algorithmic}

\end{phasebox}

\begin{phasebox}{control}{V2A model learning}
\begin{algorithmic}
\State Sample trajectories $\mathcal{B}\sim\mathcal{D}$ and timesteps $t$ from each trajectory

\State $\theta
\gets
\theta
+
\eta\nabla_\theta
\sum_{\tau\in\mathcal{B}}
\log p_{\mathtt{V2A}}
(a_t\mid o_{\leq t},o_{t+1};\theta)$
\end{algorithmic}
\end{phasebox}

\end{adjustwidth}

\end{algorithm}

\textbf{Exploration via stochastic V2A.}
Figure~\ref{fig:exploration_intuition} illustrates how this process guides exploration. Early in training, the V2A model often fails to follow the plan, so the realized next observation $o_{t+1}$ may differ substantially from the video plan $\hat{o}_{t+1}$. However, even these failed attempts provide useful training examples: the V2A model still learns to predict the action that produced the observed transition. As the model improves, the agent can follow the planner more closely and reach later stages of the task. In turn, these rollouts provide data to improve the V2A model at the newly reached stages. Repeating this cycle allows the agent to progress ever further through the task. In contrast, purely random exploration often struggles to reach these stages because they require long sequences of consistently coordinated actions. 

If the V2A model were trained only on data from its own rollouts, it would collapse to predicting the same action regardless of the observations. 
We hence make the V2A model stochastic during data collection by adding noise to its predictions. 
Stochasticity also plays a second, central role: carrying the agent past the stages it cannot yet execute (Figure~\ref{fig:single_attempt}). Where the V2A model is already competent, action noise is harmless: while the agent may drift off the currently active video plan, the video planner re-plans from the state actually reached, preventing error accumulation and allowing the agent to progress.
At a milestone the V2A model has not yet learned, such as the first time a block has to be mined, the V2A model predicts the wrong action, and the agent thus fails to produce the planned transition. The planner, however, continues re-planning, repeatedly proposing the same transition. A \textit{deterministic} agent would then be blocked at this milestone, as it would have never seen the action that would lead to this correct transition. 
The stochastic V2A model is what overcomes the roadblock: eventually, the correct random action is sampled, and the resulting transition enters the replay buffer (where it is then learned by the V2A model), enabling the agent to complete the milestone in future episodes. Random exploration, in contrast, would first have to stumble upon the entire action sequence leading to the milestone before it could even attempt the critical action, see Fig.~\ref{fig:exploration_intuition}.

\begin{figure}[t]
    \centering
    \includegraphics[width=1.00\linewidth]{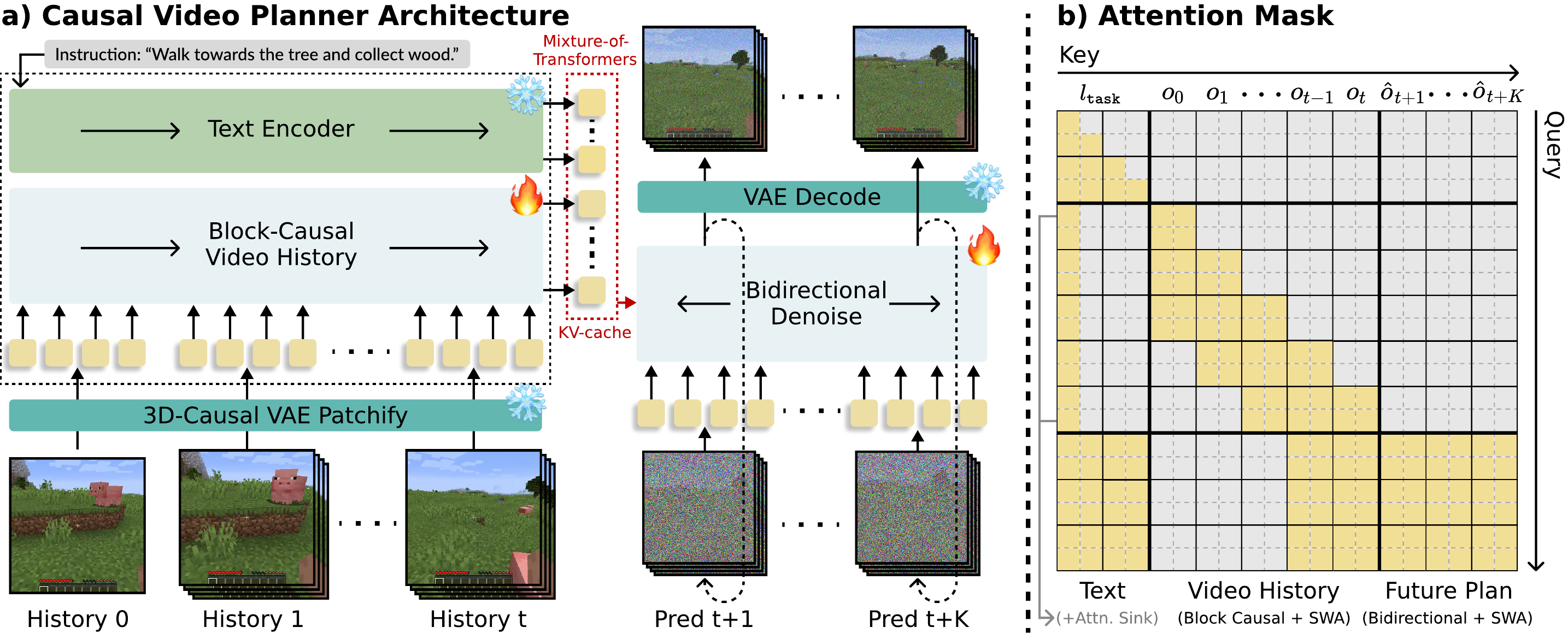}
    \vspace{-1.\baselineskip}
    \caption{
        \textbf{The causal video planner.}
        \textbf{(a)} 
        Tokenized frames and text are fed into the 14B-parameter Cosmos3-Nano~\citep{Cosmos3} model, forming a causal stream of contexts whose keys and values are cached and reused.
        The denoiser generates $K$ future frames while cross-attending to the context KV cache, so each observation and instruction is encoded only once. \textbf{(b)} Attention mask. Video history is encoded block-causally. The future plan attends to this history and bidirectionally to itself. Sliding-window attention lets the planner generalize beyond the window while keeping the KV cache within a constant budget.
    }
    \label{fig:vp_arch}
    \vspace{-1\baselineskip}
\end{figure}
\subsection{A Video Planner for Online Learning and Closed-Loop Control}
\label{sec:VP}
Following plans in closed-loop places three demands on the VP. First, it must re-plan often and cheaply. 
Second, it must run over long episodes, often well over a minute in our Minecraft benchmarks.
Third, it must stay on task even when the agent's failures push it off the distribution of expert videos.
To avoid re-encoding the history at every re-plan, we use a block-causal architecture: each observed frame is encoded once to cache its keys and values while the plan is denoised bidirectionally against this cache. Sliding-window attention over a fixed-size cache keeps the cost per plan constant, so a planner trained on 32\,s clips can run on over 5 minutes of context without degradation. 
Finally, we propose task-aware guidance, which leverages diffusion guidance~\citep{ho2022cfg,karras2024autoguidance} to guide the task-conditional video planner with a task-unconditional one, significantly improving the planner's task adherence. Figure~\ref{fig:vp_arch} and Appendix~\ref{app:VP} detail the architecture, training, and inference.

\section{Experiments}
We empirically validate \methodname{}'s ability to solve tasks across Minecraft (Section~\ref{sec:mc-results}) and traditional RL environments (Section~\ref{sec:robotics}). In Section~\ref{sec:vp-scale}, we study what emergent abilities generative video planners can provide with scale and how to leverage them. In Section~\ref{sec:robotics}, we investigate how our method generalizes to various environments and control settings. Please find video results on our \website{website.}

\subsection{Solving Long-Horizon Challenges in Minecraft}
\label{sec:mc-results}
\methodname{} solves long-horizon challenges in Minecraft without reward supervision or expert actions. On Minecraft, we pre-train a video planner on 2{,}946 hours of videos from the VPT dataset~\citep{VPT2022OpenAI}, equipping our video planner with general physical and procedural knowledge of Minecraft. The VP is finetuned further for BASALT and Minecraft survival mode. With this VP, we are able to train our self-improving V2A model from scratch with self-play.
Our agent operates in the MineRL 1.0~\citep{guss2019minerl} action space: at every step (20\,Hz), it chooses among over 4{,}000 distinct combinations of 20 keyboard buttons while moving the mouse continuously in two dimensions. We do not simplify the action space or game mechanics.

\subsubsection{Learning to Solve Fuzzy Tasks from Human Demonstration}
\label{sec:basalt}

 We first evaluate~\methodname{} on the Minecraft BASALT (Benchmark for Agents that Solve Almost-Lifelike Tasks)~\citep{Shah2021BASALT} challenge. Unlike traditional RL environments, its task objectives are intentionally fuzzy and hard-to-verify. For instance, consider the MakeWaterfall task: 

\begin{quote}
    \small\textit{Place a water bucket to make a \underline{\smash{beautiful}} waterfall. Then take an \underline{aesthetic} picture of it by moving to a good location, positioning the player to have a \underline{nice} view of the waterfall.}
\end{quote}

\begin{figure}[t]
\centering
\includegraphics[width=\linewidth]{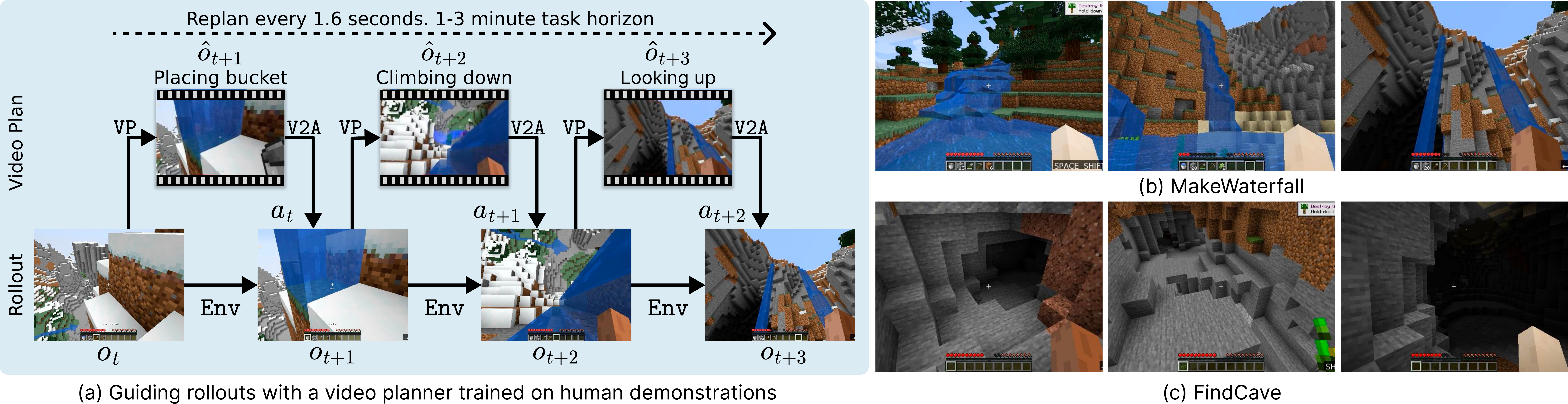}
\vspace{-2\baselineskip}
\caption{\textbf{Learning from human demonstration.} 
\methodname{} can complete a variety of open-ended BASALT tasks from self-play without any expert action labels or reward supervision.
\textbf{(a)} 
The VP inherits long-horizon planning capabilities learned from video pre-training, while the V2A model learns to ground the video plans to action via online interaction. \textbf{(b)} The constructed waterfalls are diverse and similar to human creations. \textbf{(c)} \methodname{} achieves near-human performance on FindCave.}
\label{fig:basalt}
\vspace{-1\baselineskip}
\end{figure}

 Our method is particularly suited to this kind of task where reward specification is difficult. It learns directly from fuzzy human demonstration videos without any reward supervision. 
For each task, we finetune our video planner on the corresponding BASALT human video demonstrations. We then train our V2A model from scratch only via self-play. After only 5 hours (360,000 environment steps) of in-game play, our method is already able to outperform the baselines. Figure~\ref{fig:basalt} demonstrates how \methodname{} solves these tasks. Using the same VP, we also evaluate a V2A model pretrained on expert actions from the VPT dataset.  
 
 In Table~\ref{tab:basalt}, we compare our method against an action free baseline,  VIPER~\citep{escontrela2023video}, a behavioral cloning (BC) policy fine-tuned from VPT, action-supervised methods and heuristic methods from the BASALT challenge~\citep{Milani2023BASALTRetro}, and vision-language-action (VLA) Minecraft models: STEVE-1~\citep{lifshitz2023steve1}, GROOT~\citep{cai2024groot}, JARVIS-VLA~\citep{li2025jarvis}, OpenHA~\citep{wang2025openha}. Of the BASALT challenge submissions, KAIROS is an inverse RL method based on IQ-Learn~\citep{garg2021iqlearn}, UniTeam augments the VPT BC policy with search, and GoUp executes pre-programmed scripts~\citep{Milani2023BASALTRetro}.

Even without expert actions, \methodname{} outperforms all baselines that do not rely on hand-crafted heuristics. Only 5 hours of self-play were required to achieve the highest success rate on FindCave and MakeWaterfall tasks. To the best of our knowledge, \methodname{} is the first method to achieve non-zero success on MakeWaterfall from learned behavior alone (i.e., without classifiers or hard-coded actions). 
The baselines fail in different ways (Figure~\ref{fig:waterfall}): the action-supervised methods (KAIROS, BC and UniTeam) drift out of distribution on multi-step tasks such as MakeWaterfall, the VLA models struggle to follow the task descriptions, and GoUp's hard-coded scripts are brittle, placing, for instance, the water bucket but freezing once underwater, whereas \methodname{} escapes the water using knowledge learned from human gameplay. CreateAnimalPen and BuildVillageHouse remain unsolved by all learning-based methods. Lastly, while \methodname{} does not require action labels, it can still leverage them for a further small performance gain.

\begin{table}[t]
    \caption{\textbf{Solving fuzzy tasks from BASALT.} Success rates for methods without action labels (our self-improving method and VIPER~\citep{escontrela2023video}), action-supervised and heuristic BASALT baselines~\citep{Milani2023BASALTRetro}, vision-language-action (VLA) models, and humans.}
    \label{tab:basalt}
    \centering
    \begingroup
    \fontsize{6.5}{8}\selectfont
    \setlength{\tabcolsep}{1.8pt}
    \renewcommand{\arraystretch}{1.15}

    \begin{tabular*}{\linewidth}{@{\extracolsep{\fill}}l*{13}{c}@{}}
        \toprule
        & \multicolumn{2}{c}{\textbf{No action labels}}
        & \multicolumn{4}{c}{\textbf{With action labels}}
        & \multicolumn{4}{c}{\textbf{Vision-language-action (VLA) models}}
        & \multicolumn{1}{c}{\textbf{Heuristic}}
        & \textbf{Human}\\
        \cmidrule(lr){2-3}
        \cmidrule(lr){4-7}
        \cmidrule(lr){8-11}
        \cmidrule(lr){12-12}
        \cmidrule(lr){13-13}

        {Task}
        & \begin{tabular}[c]{@{}c@{}}Mirror\\(Ours)\end{tabular}
        & {VIPER}
        & \begin{tabular}[c]{@{}c@{}}Mirror\\(Ours)\end{tabular}
        & {KAIROS}
        & \begin{tabular}[c]{@{}c@{}}BC\\(VPT-SFT)\end{tabular}
        & {UniTeam}
        & {STEVE-1}
        & {GROOT}
        & \begin{tabular}[c]{@{}c@{}}JARVIS-\\VLA\end{tabular}
        & {OpenHA}
        & {GoUp}
        & 
        \\
        \midrule
    
        {FindCave}
        & $\mathbf{82\%}$
        & 0\%
        & $\mathbf{91\%}$
        & 27\%
        & 27\%
        & 64\%
        & 45\%
        & 45\%
        & 36\%
        & 45\%
        & 36\%$^{\phantom{\dagger}}$
        & 95\%
        \\
    
        {MakeWaterfall}
        & $\mathbf{36\%}$
        & 0\%
        & $\mathbf{45\%}$
        & 0\%
        & 0\%
        & 0\%
        & 0\%
        & 0\%
        & 0\%
        & 0\%
        & \textcolor{gray}{36\%$^\dagger$}
        & 100\%
        \\
    
        {CreateAnimalPen}
        & 0\%
        & 0\%
        & 0\%
        & 0\%
        & 0\%
        & 0\%
        & 0\%
        & 0\%
        & 0\%
        & 0\%
        & \textcolor{gray}{18\%$^\dagger$}
        & 95\%
        \\
    
        {BuildVillageHouse}
        & 0\%
        & 0\%
        & 0\%
        & 0\%
        & 0\%
        & 0\%
        & 0\%
        & 0\%
        & 0\%
        & 0\%
        & \textcolor{gray}{91\%$^\dagger$}
        & 100\%
        \\
    
        \bottomrule
    \end{tabular*}

    \endgroup
    \par\vspace{2pt}
    \parbox{\linewidth}{\raggedright\scriptsize{
        $^\dagger$Obtained by pre-programmed action sequences rather than learned behavior.
    }}
\end{table}

\subsubsection{Self-improvement in Minecraft Survival Mode}
\label{sec:oracle_survival}

\begin{figure}[t]
\centering
\includegraphics[width=\linewidth]{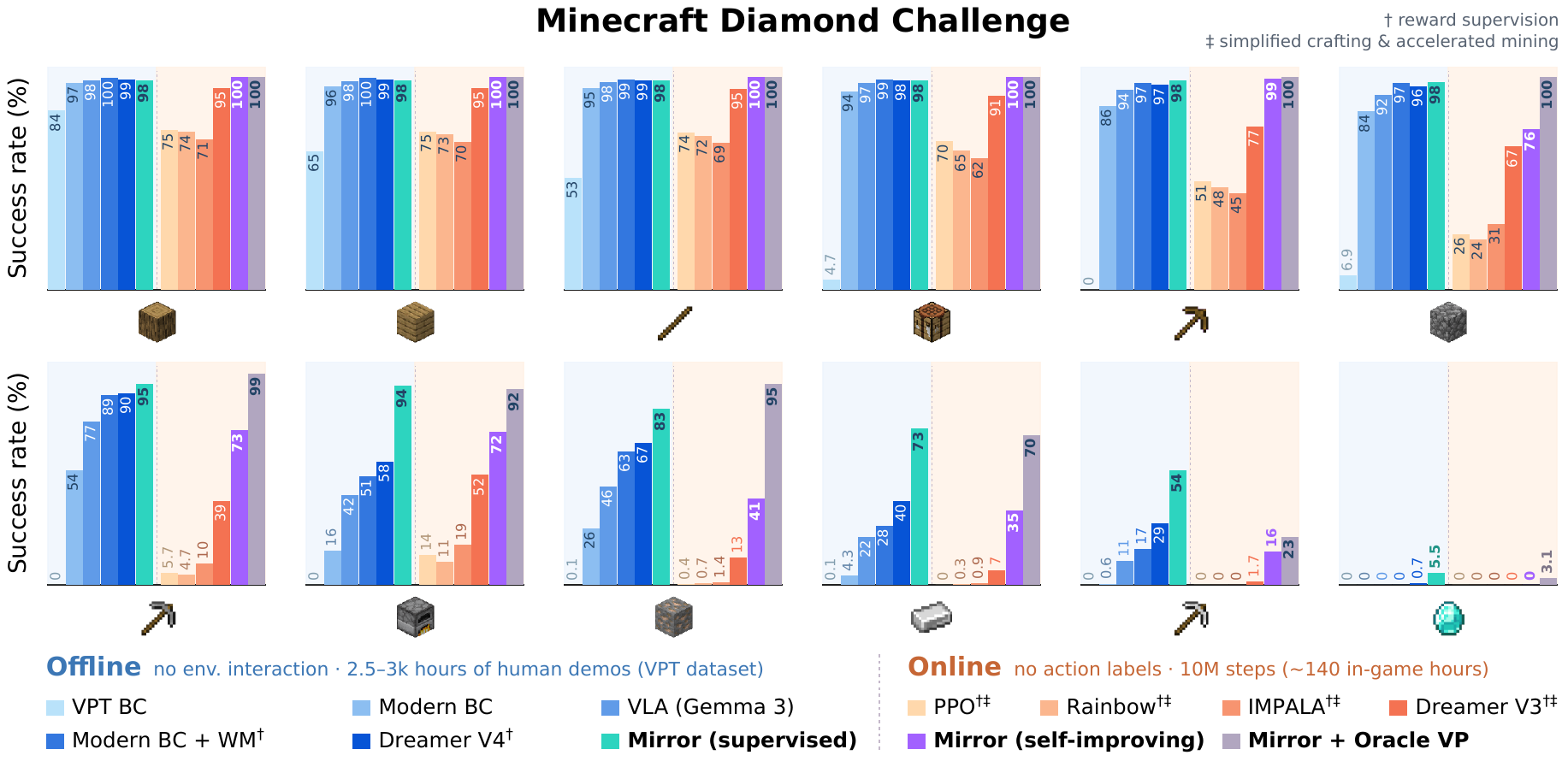}
\vspace{-1\baselineskip}
\caption{\textbf{Minecraft Diamond Challenge.} Success rates across twelve acquisition milestones. \emph{Offline} methods learn from VPT human data without environment interaction. \emph{Online} methods learn by interacting with the environment for about 10M steps (${\sim}$140 in-game hours) without expert action labels. In the offline setting, \methodname{} (supervised) sets a new state of the art. In the online setting, \methodname{} (self-improving) is the only method to reach the iron pickaxe without action labels or rewards. \methodname{} is evaluated on 128 episodes of 60 minutes with respawning, following VPT and Dreamer~V4. Offline baselines are from Dreamer~V4~\citep{hafner2025dreamerv4}. Online baselines at 10M environment steps are from Dreamer~V3~\citep{hafner2023dreamerv3}.}
\label{fig:minerl-progress}
\end{figure}

We next evaluate~\methodname{} on the Minecraft Diamond Challenge~\citep{guss2019minerl}, in which the agent must obtain a diamond in survival mode, starting from a new world and an empty inventory. In the online setting, \methodname{} is the first method to acquire an iron pickaxe through self-improvement alone, without reward supervision or expert action labels. In the offline setting, where expert action labels are available but environment interaction is not, \methodname{} obtains diamonds at nearly eight times the success rate of Dreamer~V4~\citep{hafner2025dreamerv4}, the state-of-the-art world-model agent on Minecraft (Figure~\ref{fig:minerl-progress}).

Unlike the BASALT tasks, the Diamond Challenge has an unambiguous and easily verifiable objective, but reaching it requires an extremely long horizon of up to an hour of gameplay. The agent must complete a long chain of prerequisite subgoals, from chopping trees and crafting tools to smelting iron and mining deep underground. While our video planner alone can solve the 3--5-minute BASALT tasks without any task decomposition (Sec.~\ref{sec:basalt}), an hour-long horizon remains beyond it. Such a horizon is hard even for humans: players given 20 minutes to obtain a diamond pickaxe collect diamonds in only 15\% of episodes~\citep{VPT2022OpenAI}.

To solve such an extremely long-horizon task, we follow Dreamer~V4 and condition the agent on its current subgoal.
Whereas Dreamer~V4 conditions its policy on one-hot task embeddings, we instead condition only the video planner: we finetune it on subgoal intervals of VPT human gameplay with text prompts that describe the subgoal. During rollouts, a heuristic subgoal task planner chooses the prompt from the player's own state, namely the inventory, the health, and whether a GUI screen is open. The V2A model therefore never observes the subgoal.
Since Dreamer~V4 publishes neither how it extracts subgoal intervals from the VPT recordings nor how it decides when to advance to the next subgoal during evaluation, we implement our own data curation (Appendix~\ref{app:diamond_vp_data}) and heuristic subgoal task planner (Appendix~\ref{app:task_planner}).

\paragraph{Video planning with a pretrained V2A model.} To study the video planner independently of the self-improvement algorithm, we first evaluate it with a V2A model pretrained on 3k hours of paired human action-demonstration data from the VPT dataset~\citep{VPT2022OpenAI}. In this offline setting, our agent sets a new state of the art, obtaining a diamond in 5.5\% of episodes, compared with 0.7\% for Dreamer~4~\citep{hafner2025dreamerv4}, the best world-model agent. This supports the \vpva{} decomposition in~\eqref{eqn:decompose} over \wmpi{} decompositions such as Dreamer~4's. A \wmpi{} agent must learn procedural and physical knowledge from action-labeled data, whereas the VP learns both from internet-scale video. Vision-language pretraining does not supply this knowledge either: the Gemma~3-based VLA~\citep{gemma3} evaluated by \citet{hafner2025dreamerv4} never obtains a diamond.

\paragraph{Self-improvement with a generative VP.} Without action labels or rewards, our self-improving V2A model matches its supervised counterpart on early-game items and remains competitive with offline methods trained on 2.5k hours of action-labeled demonstrations.
We begin self-improvement with a V2A model trained from scratch and a goal to acquire wood. While the V2A model may output random actions early in training, re-planning allows the agent to recover from its mistakes so it remains in distribution. Once the agent succeeds at a goal, the text prompt is updated with the next goal. The agent thus collects progressively more challenging items through online interaction. Appendix~\ref{app:diamond_v2a} details the exploration strategy and the training protocol. 
After 140 hours of online interaction (about 10M environment steps), our agent crafts a stone pickaxe in 72.7\% of episodes and an iron pickaxe in 15.6\%. At the same budget, Dreamer~V3~\citep{hafner2023dreamerv3} crafts an iron pickaxe in only 1.7\% of episodes, despite using task rewards, simplified crafting, and accelerated mining (Figure~\ref{fig:minerl-progress}). \methodname{} retains a large margin even when matching Dreamer~V3's evaluation protocol (Appendix~\ref{app:dreamerv3_protocol}), in which episodes last 30 minutes and end at the first death, rather than the 60-minute episodes with respawning used by VPT~\citep{VPT2022OpenAI} and Dreamer~V4~\citep{hafner2025dreamerv4}.

\paragraph{Self-improvement with an oracle VP.} Generative video planners are already capable, and we expect them to keep improving. To examine how \methodname{} scales with better planners, we decouple self-improvement from the generative VP using an \textit{oracle video planner} (oracle VP), which we construct from an expert policy and a copy of the environment. At each planning step, we synchronize a separate instance of the game to the agent's current state and let the expert policy, conditioned on the agent's past observations $o_{\leq t}$, play forward in it. The frames rendered by this copy serve as the plan $\hat{o}_{t+1}$, so the oracle has the same input-output behavior as a video planner, $p_{\tt{VP}}(\hat{o}_{t+1} \mid o_{\leq t})$, while the expert's actions are never shown to the V2A model.

Using VPT's early-game RL policy~\citep{VPT2022OpenAI} as the oracle, we train a V2A model from scratch with self-play. After only about 80 hours of online interaction, the agent acquires diamonds for the first time (Figure~\ref{fig:minerl-progress}), an order of magnitude fewer environment interactions than Dreamer~V3~\citep{hafner2023dreamerv3} requires, despite Dreamer~V3's use of simplified crafting and accelerated mining. RL agents take far longer because they must explore the full action space on their own. This experiment suggests that the remaining bottleneck is not \methodname{}'s ability to learn from interaction, but the capability of the video planner itself.

\subsubsection{Emergent Capabilities of the Video Planner}
\label{sec:vp-scale}

\begin{figure}[t]
    \centering
    \includegraphics[width=\linewidth]{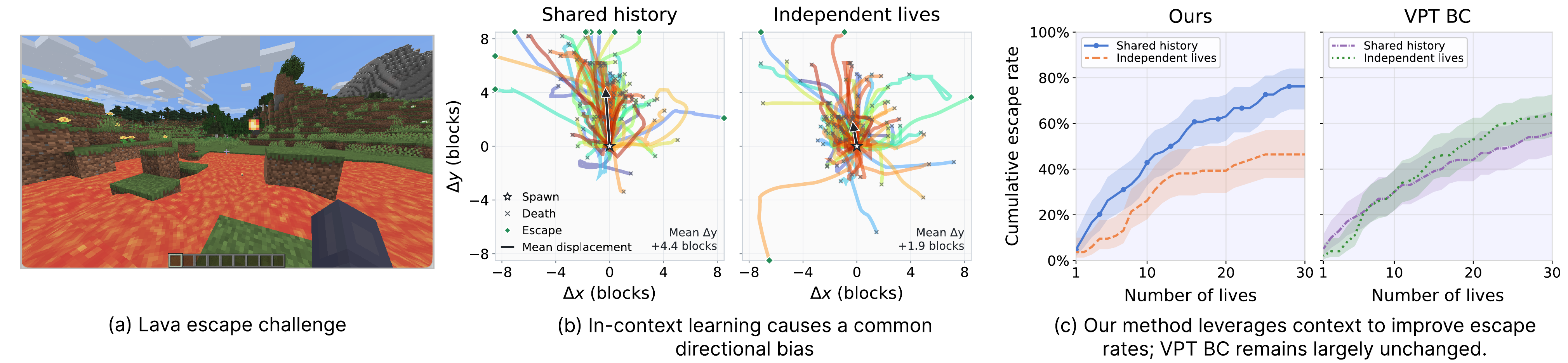}
    \vspace{-2\baselineskip}

    \caption{\textbf{Emergent in-context learning from failure.} \textbf{(a)} We challenge our agent to escape a lava lake without fine-tuning. \textbf{(b)} Retaining context between deaths causes a directional bias in the agent's trajectories, indicating in-context adaptation. \textbf{(c)} Shared history allows the agent to learn from its past failures in-context, leading to higher escape rates. Retaining recurrent state does not provide
      a comparable benefit for VPT BC.}
    \label{fig:lava-escape}
\end{figure}

\begin{figure}[t]
\centering
\begin{minipage}[c]{0.6\linewidth}
\centering
\setlength{\tabcolsep}{2.4pt}
\renewcommand{\arraystretch}{0.95}
\setlength{\aboverulesep}{0.3ex}
\setlength{\belowrulesep}{0.4ex}
\resizebox{\linewidth}{!}{%
\begin{tabular}{ll ccc cc}
\toprule
& & \multicolumn{3}{c}{\textbf{No action labels}}
  & \multicolumn{2}{c}{\textbf{With action labels}} \\
\cmidrule(lr){3-5}\cmidrule(lr){6-7}
World & Instruction
& \begin{tabular}[b]{@{}c@{}}\methodname{}\\(random)\end{tabular}
& \begin{tabular}[b]{@{}c@{}}\methodname{}\\(self-improv.)\end{tabular}
& VIPER
& BC
& \begin{tabular}[b]{@{}c@{}}\methodname{}\\(supervised)\end{tabular} \\
\midrule
\multirow{4}{*}{\rotatebox[origin=c]{90}{Woodland}}
 & Swim        & 70\%  & 80\%  & 0\% & 10\% & 80\%  \\
 & Chop tree   & 0\%   & 100\% & 0\% & 0\%  & 80\%  \\
 & Ascend hill & 100\% & 90\%  & 0\% & 0\%  & 90\%  \\
 & Dig ground  & 0\%   & 90\%  & 0\% & 0\%  & 100\% \\
\midrule
\multirow{6}{*}{\rotatebox[origin=c]{90}{Camp}}
 & Ascend hill  & 70\% & 90\%  & 0\%  & 0\%  & 90\% \\
 & Enter cave   & 90\% & 50\%  & 0\%  & 0\%  & 60\% \\
 & Light cave   & 0\%  & 0\%   & 0\%  & 0\%  & 20\% \\
 & Mine stone   & 0\%  & 10\%  & 0\%  & 0\%  & 40\% \\
 & Place blocks & 70\% & 30\%  & 20\% & 20\% & 70\% \\
 & Dig ground   & 20\% & 100\% & 0\%  & 0\%  & 80\% \\
\midrule
\multicolumn{2}{c}{Total}
 & 42\% & \textbf{64}\% & 2\% & 3\% & \textbf{71\%} \\
\bottomrule
\end{tabular}%
}
\end{minipage}\hfill%
\begin{minipage}[c]{0.4\linewidth}
\centering
\includegraphics[width=0.9\linewidth]{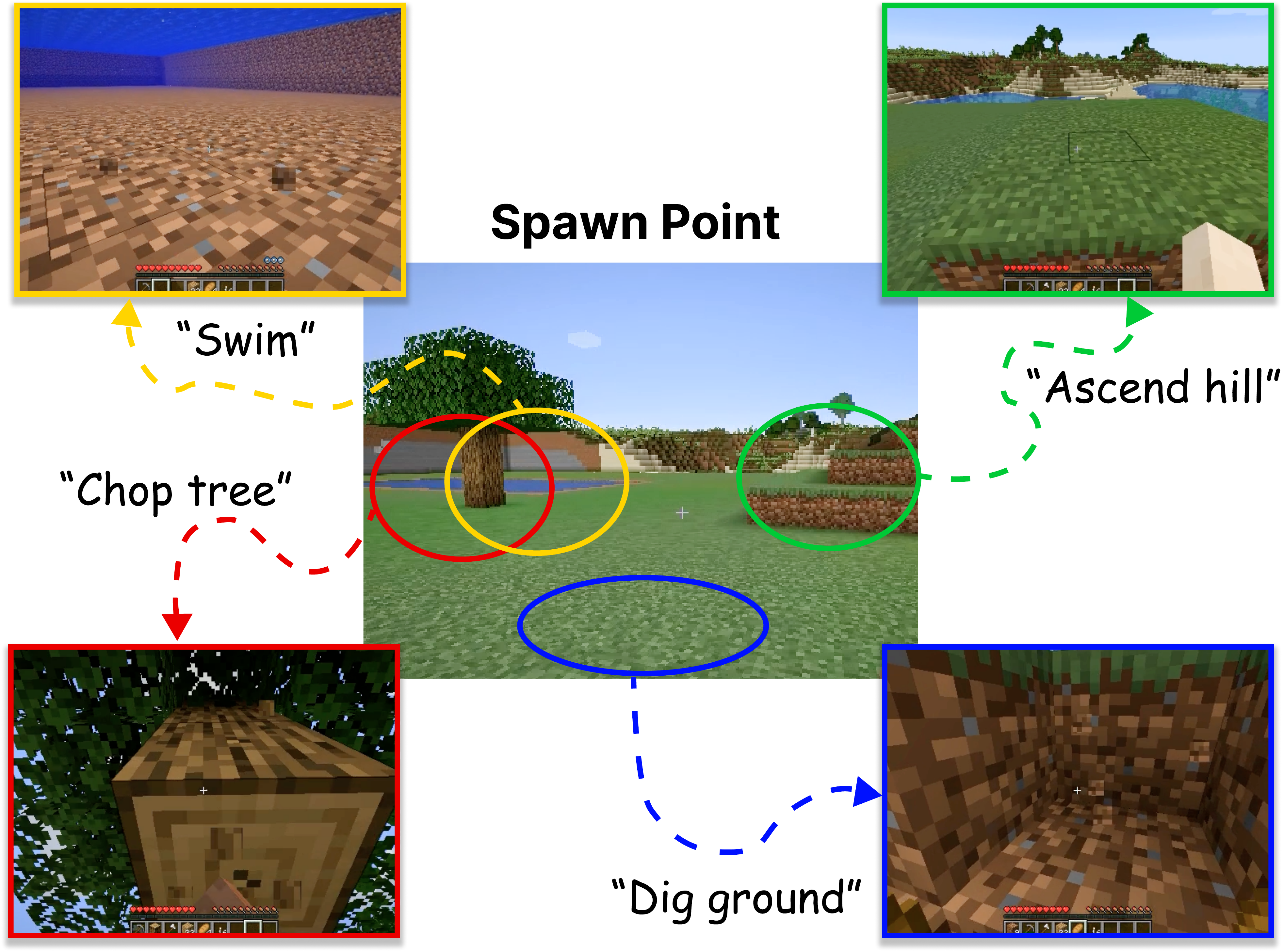}
\end{minipage}
\vspace{-0.5\baselineskip}
\caption{\textbf{Text-conditioned video planning.}
We evaluate text-conditioned control on 10 tasks across two unseen
environments (see Appendix~\ref{app:minecraft-text}).
Each task is specified only through a text prompt, with no task-specific
fine-tuning. VIPER~\citep{escontrela2023video} is included as a reward-free
video-based baseline. Combined with our self-improving V2A approach,
the agent achieves high success rates although it has never seen expert
demonstrations on these tasks.
\textit{Left:} Success rates of each method.
\textit{Right:} Text guidance leads the agent to complete different tasks.}
\label{fig:minecraft-text}
\vspace{-0.5\baselineskip}
\end{figure}

Training the video planner on internet-scale video data unlocks novel capabilities without requiring additional fine-tuning for each task. We explore such emergent behaviors in two settings.

\paragraph{Emergent in-context learning from failure.}  \label{sec:recovery}
We examine if \methodname{} can utilize its history, $o_{\leq t}$, for in-context learning. In Figure~\ref{fig:lava-escape}, we challenge our agent to escape a lava lake. If the agent falls in the lava, it respawns at the origin. We compare a case where the agent's history is preserved between deaths, and a case when the history is reset after every death. The task is difficult: with fresh histories every life, the escape rate is only 3\%. When video context is retained between lives, however, the agent leverages its past failures to improve its escape rate. Figure~\ref{fig:lava-escape} (b) shows the resulting agent's trajectories. With shared history, the agent exhibits a clear directional bias, demonstrating that prior attempts influence the agent's behavior. With independent history, the agent moves in relatively uniform directions around the starting point. Retaining history does not provide a comparable benefit to the VPT BC policy, perhaps even hindering performance. This experiment shows that our video planning agent exhibits some emergent in-context adaptation.

\paragraph{Text-conditioned video planning.} \label{sec:minecraft-text}
As a proof of concept, we test whether our pretrained video planner can be steered by text zero-shot. Since our planner was trained only on captions generated automatically by a small VLM (Appendix~\ref{app:vp_captions}), we evaluate simple goals such as ``chop down the tree'' or ``dig into the ground,'' rather than complex instructions. Accordingly, we do not aim to compete in language understanding with Minecraft-specialized vision-language-action models such as JARVIS-VLA~\citep{li2025jarvis} and OpenHA~\citep{wang2025openha}. Each task nevertheless lasts about 30 seconds and requires consistent, prolonged behavior.
We pair the text-conditioned video planner with a V2A model pre-trained on random walks, a V2A model pre-trained on the VPT expert actions, and our self-improving V2A model. We also compare against VIPER~\citep{escontrela2023video}, which similarly does not use action labels, and a text-conditioned BC policy (Appendix~\ref{app:minecraft-text}). %
As shown in Figure~\ref{fig:minecraft-text}, \methodname{} reliably harvests trees, climbs hills, and digs underground, although some tasks remain difficult across all V2A variants. The self-improving V2A model generally outperforms the one trained on random walks and remains competitive with the one supervised on human actions. 
These results show that the V2A model can execute plans without ever knowing the task. Since the V2A model is task-agnostic, a sufficiently capable video planner would let the agent solve new tasks from a text prompt alone, without task-specific action data or retraining.

\begin{figure*}[t]
  \centering
  \begin{minipage}[c]{0.615\textwidth}
    \centering
    \includegraphics[width=\linewidth]{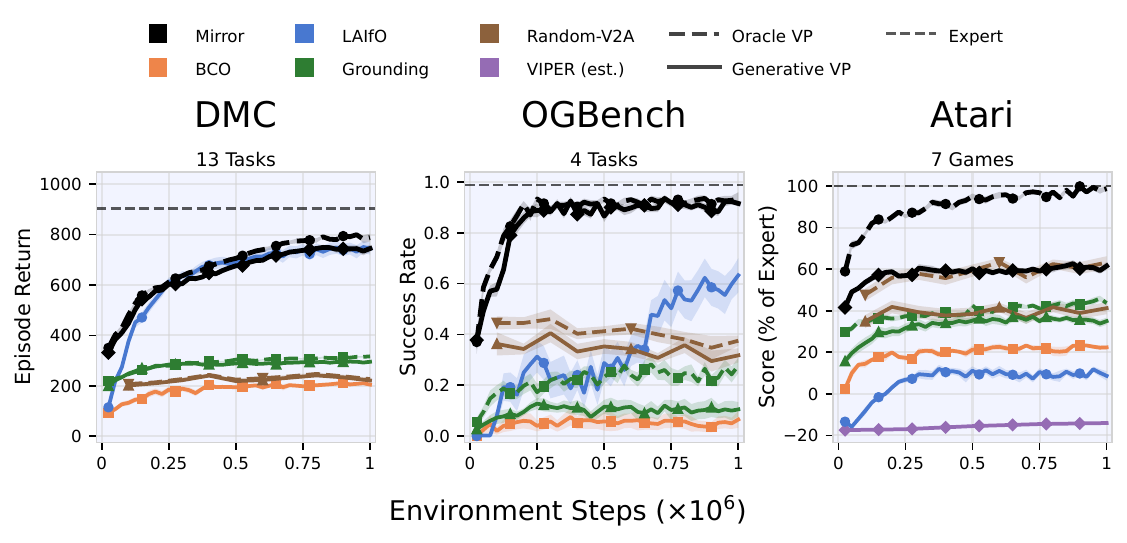}
  \end{minipage}\hfill
  \begin{minipage}[c]{0.365\textwidth}
    \centering
    \footnotesize
    \setlength{\tabcolsep}{3.5pt}
    \renewcommand{\arraystretch}{1.0}
    \resizebox{\linewidth}{!}{%
    \begin{tabular}{@{}l@{\hspace{6pt}}lccc@{}}
    \toprule
     & & \textbf{DMC} & \textbf{OGBench} & \textbf{Atari} \\
    \multicolumn{2}{@{}l}{\textbf{Final @ 1M}} & return & success & \% expert \\
    \midrule
    \textit{Oracle VP} & Mirror & \textbf{788 $\pm$ 15} & \textbf{0.93 $\pm$ 0.03} & \textbf{99 $\pm$ 1} \\
     & Grounding & 317 $\pm$ 4 & 0.28 $\pm$ 0.03 & 44 $\pm$ 1 \\
     & Random-V2A & 227 $\pm$ 5 & 0.38 $\pm$ 0.02 & 63 $\pm$ 3 \\
    \midrule
    \textit{Generative VP} & Mirror & 745 $\pm$ 15 & \textbf{0.92 $\pm$ 0.01} & 62 $\pm$ 2 \\
     & Grounding & 295 $\pm$ 3 & 0.11 $\pm$ 0.03 & 35 $\pm$ 2 \\
     & Random-V2A & 221 $\pm$ 5 & 0.32 $\pm$ 0.03 & 41 $\pm$ 3 \\
    \midrule
    \textit{Baselines} & LAIfO & 745 $\pm$ 24 & 0.63 $\pm$ 0.07 & 9 $\pm$ 2 \\
     & BCO & 205 $\pm$ 6 & 0.07 $\pm$ 0.03 & 22 $\pm$ 1 \\
    \midrule
    \multicolumn{2}{@{}l}{Expert} & 903 & 0.99 & 100 \\
    \bottomrule
    \end{tabular}

    }
  \end{minipage}
  \vspace{-4pt}
  \caption{\textbf{Imitation learning performance across three benchmark suites.}
  \textit{Left:} learning curves on 13 DeepMind Control (DMC) tasks,
  four OGBench tasks, and seven Atari games. Colors identify methods;
  dashed method curves use an oracle VP and solid method curves use a generative VP.
  The horizontal dashed line marks expert performance.
  Curves average tasks equally, with shaded regions showing $\pm 1$ standard error
  across three seeds per task.
  The Atari VIPER curve is digitized from \citet{escontrela2023video}
  over its first 1M decisions and rescaled to \% of expert.
  \textit{Right:} final performance at 1M environment transitions
  (mean $\pm$ SE, equal task weights), separated by planner type.\looseness=-1}
  \label{fig:three_suites}
  \vspace{-6pt}
\end{figure*}

\subsection{Solving Robotics Tasks and Video Games}
\label{sec:robotics}

\textbf{Setup.} We evaluate Mirror on standard subsets of key synthetic decision-making benchmarks: OGBench~\citep{park2025ogbench}, DeepMindControl~\citep{tassa2018deepmindcontrolsuite}, and Atari~\citep{bellemare2013atari,escontrela2023video}.
We benchmark with BCO~\citep{torabi2018behavioralcloningobservation}, which alternates between learning a V2A on policy rollouts and cloning its inferred actions on expert demonstrations, VIPER~\citep{escontrela2023video}, which uses video prediction likelihoods as rewards, LAIfO~\citep{giammarino2024adversarialimitationlearningvisual}, which learns a policy by adversarially matching expert and agent transitions, and the method proposed by \cite{Luo2024GroundingVM} (which we abbreviate as ``Grounding'') which learns a goal-conditioned policy to reach video-predicted subgoals. 
We evaluate \methodname{} with both a learned video planner and an oracle planner that supplies future frames from an RL expert. Unlike in our Minecraft setting, large-scale human play datasets are not available, and we thus found it essential to augment the generative planner's training data with recovery examples collected using DAgger~\citep{ross2011reductionimitationlearningstructured}.
Please find an in-depth discussion of dataset construction, evaluated tasks, planner training, and baseline implementations in Appendix \ref{app:baseline_details}.

\textbf{Generative VP results.} As shown in Figure~\ref{fig:three_suites}, among methods evaluated without oracle planners, \methodname{} surpasses all baselines on Atari and OGBench, and learns faster than LAIfO on DMC. LAIfO performs less consistently across domains, learning less efficiently on OGBench and achieving limited returns on Atari. VIPER also achieves low returns on Atari, suggesting that using video model predictions directly provides better supervision than likelihood-based rewards. 

The comparisons with BCO and Grounding help explain these gains. Although BCO also learns a V2A model from self-collected experience, it performs substantially worse, indicating the importance of using the video planner to guide data collection. Grounding improves on BCO, consistent with the benefit of video-guided exploration, but still remains substantially below \methodname{}. The remaining gap suggests an advantage to learning a V2A model that translates frame-to-frame transitions into actions over learning a goal-conditioned policy to reach predicted subgoals. 

\textbf{Oracle VP results.} To evaluate our self-improvement algorithm independently of video-planner quality, we also use an Oracle VP that supplies frames from an RL expert. In this setting, our method matches the expert on Atari, suggesting that our Atari performance is limited by the video planner, most likely by the compute needed to train a model that accurately predicts fine details such as the motion of small projectiles. This result demonstrates that as video models continue to scale and improve, our method provides a simple framework that is able to solve a wide variety of tasks. In Appendix \ref{app:ogbench_generalization_analysis}, we also demonstrate that \methodname{} learns a V2A model that generalizes better to unseen goals. 

\methodname{} also exceeds the expert's return in Assault and Zaxxon by 10\% and 25\%, respectively. We observe that the expert issues redundant fire commands that do not produce additional shots. The V2A model can thus follow the plan without copying them. As a result, \methodname{} makes fewer ineffective firing attempts in Assault, reducing overheating, and releases FIRE more frequently in Zaxxon, enabling subsequent presses to trigger new shots.

\section{Conclusion}
We have introduced \methodname{}, a method that learns to translate behaviors encoded in large video planners to policies executable by an agent. Unlike prior work, our method assumes neither paired video-action data nor task-specific rewards, but instead learns to translate video into actions simply by self-play in an interactive environment.

\paragraph{Limitations.} Our approach has several limitations. First, without a separate high-level task planner that conditions the video planner on subgoals, our agent could not cover the hour-long horizon of the diamond task. We expect capable VLM-based task planners to fill this role in the future: VLMs excel at abstract, high-level task planning but struggle with dense, low-level control, whereas video planners show the opposite strengths, making the two naturally complementary.
Second, with self-improvement alone, our agent reaches the iron pickaxe but does not yet mine diamonds. An imperfect V2A model often leads the agent into suboptimal states that human players rarely visit, and recovering from them demands very strong generalization from the video planner. Our video planner falls short of this at later milestones, each of which is covered by only a few hours of human gameplay in our training data.
Third, some actions leave no visible trace, so inferring them from vision alone and from scratch is ill-posed; in practice, we expect a small amount of action data to resolve this. 
Lastly, training a capable generative video planner can be data-intensive. Unlike human gameplay, which naturally contains mistakes and recoveries, the data generated by expert policies in our robotics and control benchmarks (Section~\ref{sec:robotics}) lacked diversity, so we needed to augment it with recovery data collected via DAgger~\citep{ross2011reductionimitationlearningstructured} to train a sufficiently robust video planner.

We believe \methodname{} offers a promising path toward general-purpose agents that acquire diverse behaviors from video pre-training and learn to self-improve their execution through interaction.

\paragraph{Acknowledgements.} This work was supported by the Toyota Research Institute (TRI) University 3.0 (URP) program, the National Science Foundation under Grant No. 2211260 and Cooperative Agreement EEC-2330040 (NSF Engineering Research Center for Human AugmentatioN via Dexterity, HAND), AMD via the MIT AI Hardware Program, and the Amazon Science Hub. We further gratefully acknowledge significant compute and technical support by AMD. The views and conclusions contained in this document are those of the authors and should not be interpreted as necessarily representing the official policies, either expressed or implied, of any other entity. We thank Chonghyuk Song for helpful discussions.

\paragraph{AI Use Statement.}
We used generative AI tools to assist with the implementation and editing of experimental code, literature retrieval and discovery, the generation of synthetic data, and language polishing of the manuscript for clarity and readability. Generative AI tools were not used to formulate or prove mathematical claims. All experimental results and conclusions were independently evaluated by the authors. The authors take full responsibility for the final content of this work.

\bibliography{iclr2027_conference}
\bibliographystyle{iclr2027_conference}

\clearpage
\appendix
\section{Video Planner}
\label{app:VP}
\paragraph{Design principle.} 
The central goal behind the design of our video planner is long context. Many of our tasks require memory over long horizons. Long context may also help error recovery in closed-loop planning. Early in training, the imperfect V2A model often deviates from expert behavior, forcing the planner to operate on trajectories outside its training distribution. With a short context, we often saw the planner repeat the same failed behavior. With long context, we instead observed emergent recovery behavior, as the planner can use its previous failures to change strategy (see Sec.~\ref{sec:recovery}) and replan. This section describes how we make such long context affordable.

\subsection{Architecture}
Videos are encoded with the Wan2.2 VAE~\citep{wan2025wan}, which compresses 4 frames in time and $16\times16$ pixels in space into 48 latent channels. The zeroth pixel frame is encoded on its own rather than grouped with other frames, so $L$ latent frames correspond to $1+4(L-1)$ pixel frames. The MineRL environment~\citep{VPT2022OpenAI} renders $640\times360$ frames at 20\,fps. We resize each frame to $320\times192$, so that every four-frame chunk (0.2\,s) is encoded into a $12\times20\times48$ latent array (height $\times$ width $\times$ channels). A $2\times2$ patchifier then turns each latent frame into 60 tokens of 192 dimensions, which form the input of the video planner.

\paragraph{Backbone.}
The planner is initialized from Cosmos3-Nano~\citep{Cosmos3}, a mixture of transformers~\citep{liang2024mixture} with a text tower and a video tower. The two towers share the same Qwen3-VL \citep{bai2025qwen3} block design but have separate weights. Each has 7B parameters, 36 layers of width 4096, 32q\,/\,8kv GQA heads of dimension 128, SwiGLU MLPs of width 12{,}288. For positional embedding, 3D-MROPE~\citep{su2024roformer,bai2025qwen3,Cosmos3} is used in which the $128$-dim attention head is split into $[48, 24, 24]$-dims, each encoding the time, height and width of the video. We freeze the text tower and finetune only the video tower. The video tower plays two roles: as a \emph{context encoder}, it encodes the clean history of observed frames, and as a \emph{denoiser}, it denoises the frames of the plan. We refer to the two roles separately for clarity, although they share the same architecture and weights.

\paragraph{Block-sparse attention with FlexAttention.}
The original Cosmos3 implementation uses NATTEN~\citep{hassani2023neighborhood}, which cannot efficiently express our block-causal sliding-window attention pattern. We instead use FlexAttention~\citep{dong2024flex}, which supports efficient attention with arbitrary block-sparse masks. On NVIDIA Hopper and Blackwell GPUs, we use FlexAttention's experimental FlashAttention-4~\citep{zadouri2026flashattention} backend, which is 1.35--1.55$\times$ faster than the default Triton~\citep{tillet2019triton} backend on our workload. On AMD GPUs, we use the Triton backend.

\paragraph{Mixture-of-transformers attention.}
In the original Cosmos3, all video tokens attend to each other bidirectionally under a single mask. In our implementation, however, the video tower's role is split into a context encoder and a denoiser, which require distinct attention patterns. The video history is block-causally encoded frame by frame, whereas the future plans are jointly denoised and thus require bidirectional attention. Together with the text, this yields three token streams. At every layer, each stream self-attends to its own tokens and cross-attends to the streams it is conditioned on:
\begin{itemize}[leftmargin=20pt,itemsep=0pt,topsep=2pt]
    \item \textbf{Text} (text tower): self-attends causally, token by token. It carries the video metadata (resolution, aspect ratio, duration and frame rate), which also serves as an attention sink, and optional task instructions.
    \item \textbf{History} (context encoder): self-attends block-causally, i.e., bidirectionally within a frame and causally across frames. Cross-attends to the text.
    \item \textbf{Plan} (denoiser): self-attends bidirectionally within the plan. Cross-attends to the history and the text.
\end{itemize}
Handling all these attention patterns in a single joint attention would require an overly complex attention mask. This is not only cumbersome to implement but also substantially slows down the FlexAttention kernel. We instead compute attention with a simpler mask for each source and merge the results as
\begin{equation}
o=\sum_{s}\frac{\exp(\mathrm{LSE}_s)}{\sum_{s'}\exp(\mathrm{LSE}_{s'})}\,o_s ,
\end{equation}
where the attention output $o_s$ and the log-sum-exp $\mathrm{LSE}_s$ are computed separately for each source $s$. This merge is mathematically equivalent to standard scaled dot-product attention (SDPA) computed jointly over all sources. Keeping the sources separate makes the implementation of variable-length training and cached inference straightforward, at negligible overhead.

\paragraph{Attention sink.}
In video models such as Wan~\citep{wan2025wan}, the zeroth latent frame serves as the attention sink. In Cosmos3, whose mixture-of-transformers structure lets every video token attend to the text, the text tokens appear to play this role instead. Removing the text tokens greatly slowed down the finetuning. Provided that every video query can properly attend to the text metadata tokens, adapting the pretrained bidirectional model to block-causal attention is fast, taking fewer than 1,000 finetuning steps. On the other hand, removing the zeroth latent frame from the attention context of the denoiser caused only a small drop in performance if the history is long enough. The sliding-window attention described below therefore does not keep the zeroth latent frame; instead, every query always attends to the text.

\paragraph{Text conditioning format.}
Our video planner is conditioned on two kinds of captions, written in Cosmos3's structured JSON format and wrapped in a chat template. Each caption contains the video metadata (resolution, aspect ratio, duration and frame rate) and, optionally, temporal information and language instructions.\looseness=-1

The context encoder takes a \emph{task-level caption}, which describes the long-horizon task sequence. It can optionally state how much time has elapsed since the current task sequence began. We find that providing this resume timestamp helps when the model context must be reset in the middle of a task. The caption may also describe the preceding clip, although we do not use this in our experiments. With all fields present, it reads:
\begin{lstlisting}[style=appcode]
{"temporal_caption": "<observations of the preceding 10 s> The clip resumes at 1:23.4. Task sequence: collect wood; then craft a crafting table", "resolution": {"H": 192, "W": 320}, "aspect_ratio": "5,3", "duration": "32.0s", "fps": 20}
\end{lstlisting}
The denoiser takes a \emph{control-level caption}, which describes the desired short-horizon behavior of the plan. The \texttt{duration} field is included to help the denoiser handle variable-length prediction. It can optionally state the time range of the plan to generate. An unconditional plan uses the metadata-only caption.
\begin{lstlisting}[style=appcode]
{"actions": [{"time": "1:25.0-1:26.6", "description": "Place the crafting table"}], "resolution": {"H": 192, "W": 320}, "aspect_ratio": "5,3", "duration": "1.6s", "fps": 20}
\end{lstlisting}

\subsection{Training}
\paragraph{Objective.}
We train with the linear interpolant $x_\tau=(1-\tau)x+\tau\epsilon$ and velocity prediction $v=\epsilon-x$~\citep{lipman2022flow,liu2022flow}, using a squared error averaged over latent dimensions. Every target latent frame carries equal weight in the loss. Following~\citet{esser2024sd3} and Wan~\citep{wan2025wan}, $\tau$ is drawn from a mixture of a logit-normal distribution shifted towards higher noise scale. As our plan resolution and length are smaller than those of typical video models, we use relatively low shift. We use $\mathtt{shift}=2$ for VPT pretraining and BASALT finetuning, and use $3$ for the diamond finetuning.
We also sample $\tau$ from a uniform distribution 10\% of the time to ensure better GUI interaction details and likelihood estimation.

\paragraph{Variable plan length.}
The right length for a plan is unknown in advance. Using too short plans or planning autoregressively with the shortest chunk the VAE allows (one latent frame, 0.2\,s) produced poor plans, while 4\,s is too long for open-loop execution. We therefore train a single model on a mixture of plan lengths from 0.2\,s to 4\,s. All frames of a prediction window share one diffusion time, and different windows draw diffusion time and noise independently. We also stochastically drop the text conditions of a subset of the prediction windows for unconditional generation and text-conditional classifier-free guidance~\citep{ho2022cfg}.

Each training sample is a 32\,s video clip, which is block-causally encoded as the clean history. Because this history is shared, we predict many prediction windows from a single video clip at once.
This reuse of the context encoding is our key scaling knob. The learning signal comes only from the denoised tokens, so predicting many windows per video clip maximizes the learning signal per token. In our training setup, the prediction windows of each video clip span twice its 32\,s length, totaling 64\,s. Since the windows differ in position, length, diffusion time and noise, predicting the same video clip twice still provides new learning signal. In our gradient noise scale~\citep{mccandlish2018largebatch} measurement at 30k steps, this double pass over a video clip reduced the gradient noise as much as an additional fresh video clip would. As the cost of encoding 32\,s of history is amortized over all windows of the video clip, long-context training becomes affordable. Combined with sliding-window attention, this long-context training lets the planner roll out far beyond the training length, to about five minutes in our experiments.
\begin{figure}[t]
\centering
\includegraphics[width=\linewidth]{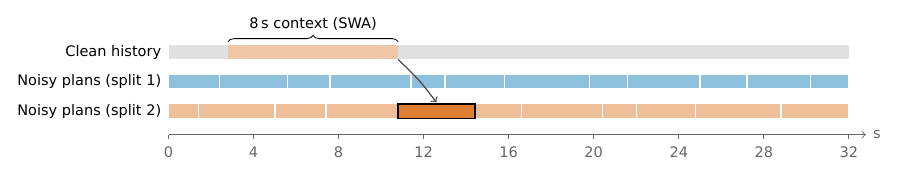}
\vspace{-2\baselineskip}
\caption{\textbf{Variable prediction length training.} Prediction windows of variable length over the same 32\,s clip. The clean history is encoded once by the context encoder and shared by all prediction windows. With sliding-window attention, the clean history tokens and the noisy plan windows attend to the preceding 12\,s and 8\,s clean frames, respectively.}
\label{fig:vp_windows}
\end{figure}

\paragraph{Static shape training for variable-length plans.}
Compiling every transformer block is essential for training speed, and it requires static tensor shapes. FlexAttention makes this even more important: its block mask is built for a specific sequence shape, so any change of shape forces the mask and the attention kernel to be rebuilt and recompiled. While FlexAttention also supports dynamic-shape compilation, it is far less efficient than static compilation in both throughput and memory. We therefore keep every tensor shape fixed for training and express the variable prediction windows purely through the attention mask. We illustrate this in Figure~\ref{fig:vp_windows}.

In our setup, each GPU processes 6{,}144 text tokens for the frozen text tower and one 32\,s video clip as a fixed-length sequence of 9{,}600 clean video tokens (160 latent frames $\times$ 60 tokens), followed by 19{,}200 noisy plan tokens. All three numbers are multiples of FlexAttention's 128-token block size, so the sequences align exactly with the $128\times128$ attention blocks for best efficiency. The number, lengths and positions of the prediction windows change at every step, but they only change the values of the boolean and integer window/frame id tensors that the mask function reads. The mask function for the video self-attention is:
\begin{align*}
\mathtt{attend}(q,k)
= \big[\,k \text{ is clean} \;\wedge\; k \in \mathtt{SWA}(q) \,\big] \;\vee\; \big[\,q, k \text{ are noisy} \;\wedge\; \mathtt{plan\_id}(q)=\mathtt{plan\_id}(k)\,\big],
\end{align*}
where $\mathtt{SWA}(q)$ is the set of clean tokens in the sliding window of $q$. For a clean query, this is the preceding 12\,s up to and including its own frame. For a noisy query, this corresponds to the 8\,s of clean frames immediately before its prediction window. $\mathtt{plan\_id}(q)$ is the index of the prediction window containing $q$. We use $\mathtt{plan\_id}(q)=0$ for clean tokens.

\paragraph{Parallelism and memory.}
We shard the video tower with Hybrid Sharded Data Parallel (HSDP) strategy on multiple nodes using Pytorch FSDP2 library~\citep{zhao2023fsdp}. Parameters are sharded within each 8-GPU node and replicated across nodes. We use activation checkpointing~\citep{chen2016checkpoint} for each transformer block. On AMD MI350X GPUs (288\,GB of memory) we do not use activation checkpointing.

\paragraph{Precision.}
We use BF16 precision for the computation. FSDP2 automatically downcasts the FP32 weights to BF16. The gradients are reduced across ranks in FP32. Numerically sensitive parts stay in FP32: the input patch projection, the timestep embedder, the velocity head, RMS/LayerNorm, the rotary embeddings, and the LSE attention merge. Training losses are computed in double precision. The frozen text tower and the VAE run in BF16. 
For pretraining, we additionally use FP8 mixed precision training with dynamic tensorwise scaling for the linear layers.
Because we use sliding window attention (SWA), which is computed by an efficient block-sparse kernel, the training step is dominated by these matrix multiplications: FP8 speeds them up by about 1.6$\times$ and a full training step by about 1.3$\times$. We found no evidence that FP8 training degrades the planning quality. FP8 can be disabled for cooldown or finetuning to reach a lower loss floor. 

\paragraph{Optimization and pretraining schedule.}
We use AdamW~\citep{loshchilov2019adamw} with $\beta=(0.9,0.95)$, $\epsilon=10^{-8}$ and weight decay 0.01, a 1{,}000-step linear warm-up followed by a constant learning rate, gradient clipping at global norm 3.0, and an EMA of the weights with decay 0.9999. We chose AdamW because Cosmos3 was trained with it. We pretrained on VPT contractor dataset~\citep{VPT2022OpenAI} for 220k steps. As GPUs became available we grew the global batch from 8 to 32, raising the learning rate roughly with the square root of the batch size (Table~\ref{tab:vp_schedule}), and ended with a 5k-step cooldown at a constant $5\times10^{-6}$. The run took roughly 4{,}700 H200 GPU-hours of training steps. The final pretrain checkpoint is the step-220k EMA.

\begin{table}[h]
\centering
\small
\caption{Pretraining schedule of the video planner on VPT dataset.}
\label{tab:vp_schedule}
\begin{tabular}{lcccc}
\toprule
Steps & GPUs & Global batch & Learning rate & Wall-clock \\
\midrule
0--30k & 8 H200 & 8 & $1.5\times10^{-5}$ & 25\,h \\
30k--80k & 16 H200 (2 nodes) & 16 & $2.0\times10^{-5}$ & 42\,h \\
80k--215k & 32 H200 (4 nodes) & 32 & $2.7\times10^{-5}$ & 115\,h\\
215k--220k & 8 H200 & 32 & $5\times10^{-6}$ (cooldown) & 10\,h \\
\bottomrule
\end{tabular}
\end{table}

\paragraph{Pretraining data.}
The VPT contractor dataset contains 2{,}945.7 hours of gameplay. We excluded one game-version subset (6.9; 271 recordings, 21.8\,h) whose logged cursor coordinates are malformed; the coordinates are needed because the recordings do not show the cursor inside menus, so we draw it back into those frames. We also excluded low-quality data (7.9\,h), leaving 55{,}919 recordings (2{,}916.0\,h) for training. Training samples a recording uniformly and then a random 32\,s clip from it. Frames are resized from $640\times360$ to $320\times192$ with bilinear interpolation and antialiasing. Over 220k steps, the planner saw about 98{,}000 hours of noisy video clips. The captions are generated from the videos with the Cosmos3-Nano VLM, as described in Appendix~\ref{app:vp_captions}.

\paragraph{Finetuning.}
The pretrained video planner is finetuned for each downstream task with global batch 8 (16 to 32 for the Diamond planners) and an EMA decay of 0.9995. The number of steps varies from 10k to 50k depending on the task difficulty and the amount of available video data. The BASALT planners are trained only on human gameplay videos, without text instructions, with a learning rate of $10^{-5}$, a 1k-step warm-up and cosine decay. BASALT episodes end when the player presses the Esc button, which is not visible in the video. We therefore finetune the video planner to predict a gray screen once the player presses the Esc button. The Diamond planners were finetuned with 16 MI350X GPUs on our curated split of human VPT gameplay, divided into subgoal clips (Appendix~\ref{app:diamond_vp_data}). Text conditioning informs the planner of the current stage of the diamond tech tree.

\subsection{Inference}
\label{app:vp_inference}

\begin{figure}[t]
\centering
\includegraphics[width=\linewidth]{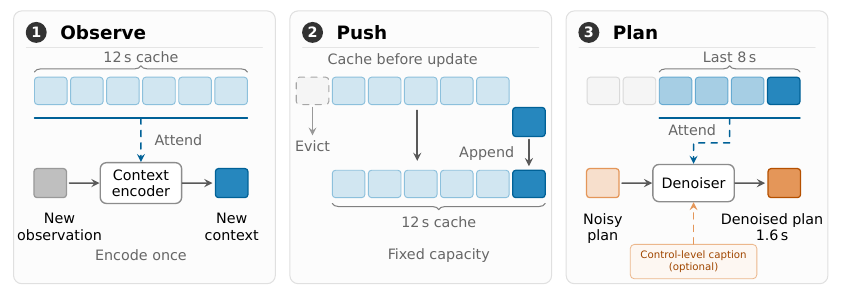}
\vspace{-2.5\baselineskip}
\caption{\textbf{Rolling cache at inference.}
(1) Each new observation is encoded once by the context encoder,
which attends to the 12\,s cache.
(2) The resulting keys and values are appended and the oldest context
is evicted, keeping the cache size fixed.
(3) The denoiser generates the next 1.6\,s of video while attending
to the last 8\,s of cached context and, optionally, to a control-level
caption. Cache cells schematically group frames.}
\label{fig:vp_cache}
\end{figure}

\paragraph{Rolling cache with a fixed budget.}
At inference, the planner keeps the keys and values of only the most recent 12\,s of observations in a fixed-size ring-buffer cache. Each newly observed frame is encoded once by the context encoder using this cache. This new frame replaces the oldest frame in the cache, which no query can attend to anymore because of the sliding window. To plan, the denoiser generates the next 1.6\,s future by attending to the last 8\,s of the cache and optionally to the control-level caption. Because the cache size never grows, every plan costs the same and the model can be compiled with static shape. We illustrate this in Figure~\ref{fig:vp_cache}.

\paragraph{Sampling.}
Plans are sampled with the second-order UniPC solver~\citep{zhao2023unipc} for 20 steps with a time shift of 3. We quantize the linear layers of the video tower to FP8 with per-tensor dynamic scaling, which makes planning $1.43\times$ faster than BF16 on an H200. With 20 solver steps and autoguidance, the planner generates about 4.2\,s of video per second of computation on a single H200.

\paragraph{Task-aware guidance.}
We use two forms of task-aware guidance. For the BASALT tasks, we use a variant of autoguidance~\citep{karras2024autoguidance}, where the positive model is the finetuned planner and the negative model is a planner trained on broader, less task-specific data, such as its parent before task finetuning; we use a guidance scale of 1.5. For Sec.~\ref{sec:minecraft-text} and Sec.~\ref{sec:oracle_survival}, we use text classifier-free guidance~\citep{ho2022cfg} with a scale of 7. For Sec.~\ref{sec:oracle_survival}, the text consists of one of 16 subgoal captions from log to diamond, chosen by the subgoal task planner (Appendix~\ref{app:task_planner}). The conditional and unconditional branches share the cached history, so guidance doubles only the denoiser's computation. We find that task-aware guidance is crucial for goal persistence. Without it, the planner often abandons the current goal after repeated failures. As a result, it may skip necessary steps and move on to the next subgoal before completing the current one.
Task-aware classifier-free guidance effectively suppresses repeated mistakes and strengthens adherence to the intended goal.

\subsection{Caption Generation}
\label{app:vp_captions}

\paragraph{Pipeline.}
We captioned the VPT corpus with the vision-language model of Cosmos3-Nano~\citep{Cosmos3}, served offline by vLLM~\citep{kwon2023vllm} with JSON-schema-constrained decoding (temperature 0.7, top-$p$ 0.8, top-$k$ 20). We aim for text that is both \emph{descriptive}, grounded in what is visible, and \emph{instructional}, phrased the way a user would command an agent. We therefore generate captions in three stages. A strictly factual stage comes first, and later stages only rephrase or group its output.
\begin{enumerate}[leftmargin=*,itemsep=1pt,topsep=2pt]
    \item \emph{Descriptions.} The model watches overlapping windows of every video at four temporal scales. These are 8\,s windows every 4\,s at 4\,fps, 24\,s windows every 12\,s at 1\,fps, and 64\,s windows every 32\,s at 0.5\,fps, all at $192\times320$. The fourth scale is short windows at full $640\times360$ resolution centered on salient moments (attacks, opening a menu, jumps, sharp turns, climbs and descents of at least four blocks). The model is not told why a window was chosen. The model lists up to six atomic, intent-free descriptions per window, each with a start and end time.
    \item \emph{Instructions.} Batches of 40 descriptions, together with the video covering their time span, are rewritten one-to-one into imperative instructions. Each instruction inherits the description's time span.
    \item \emph{Tasks.} From the video duration and up to 110 descriptions evenly distributed across time, the model partitions the whole recording into 1--8 consecutive task segments and infers the intent that makes each segment's behavior coherent. We run this stage five times per video with different seeds, which yields five alternative task partitions.
\end{enumerate}
The description stage took 4.6 hours on 32 H200 GPUs, plus 1.2 hours for the coarsest scale.

\paragraph{Prompts.}
The instructions of the three stages, verbatim:
\begin{lstlisting}[style=appcode,title={\scriptsize\textbf{Stage 1: descriptions} (system: ``You are a helpful assistant.''; input: video)}]
This is a clip from a Minecraft gameplay recording; each frame is preceded by its <t seconds> timestamp within the full episode. List the distinct things the player OBSERVABLY does, as separate facts. For each fact give start_s and end_s in those same episode seconds, read from the frame timestamps. Rules: describe only what is visible on screen — never goals, purposes or attempts (no 'trying to', 'in order to', 'prepares to', 'searching for'); one atomic action or event per fact; repeated actions separated in time are separate facts; name blocks/items/mobs/GUI screens concretely when identifiable (hotbar, inventory, crafting table, crosshair target); 6-20 words per fact. If nothing notable happens, return one fact describing the ongoing state (movement direction, terrain).
\end{lstlisting}
\begin{lstlisting}[style=appcode,title={\scriptsize\textbf{Stage 2: instructions} (system: ``You are a precise Minecraft gameplay annotator.''; input: video + indexed descriptions)}]
Below are OBSERVATIONS of what a Minecraft player did, each on its own line with an index. Rewrite EVERY observation as a CONTROL: the instruction a coach could have given just before it, which that behaviour then obeys.
Rules for each control:
- Imperative, 3-14 words, concrete Minecraft vocabulary. Command the ACTION, not the scene: 'Mine the coal seam in the left wall', not 'The player mines coal'.
- Say only what the observation says. Add no goals, no reasons, no purpose clauses ('to get wood', 'so that...'), no outcomes the player could not know in advance.
- A coach speaks before the fact, so drop discoveries and results from the wording: 'Open the furnace' rather than 'Open the furnace, revealing an iron ingot'.
- Keep the affirmative form; never write 'Don't', 'Avoid' or 'Stop'.
- Return one control per index, for every index given, in the same order.
\end{lstlisting}
\begin{lstlisting}[style=appcode,title={\scriptsize\textbf{Stage 3: tasks} (system: ``You are a precise annotator of MINECRAFT gameplay. Every session, task and control below comes from a recording of a human playing Minecraft, and every word you write must be Minecraft vocabulary.''; input: text only)}]
You are splitting one Minecraft recording into its TASKS.

A TASK answers exactly one question:

    What ongoing purpose makes this stretch of the trajectory coherent?

It is NOT a summary of what happened in order.

  BAD  (activity summary): "Walks around, gathers wood, opens the inventory, and mines."
  GOOD (intent hypothesis): "Prepare equipment for underground exploration."

The segments run back to back and together they cover the WHOLE recording, from the
first frame to the last. You give only the END of each segment: the first starts at
0 s, every later one starts where the previous ended, and the last one ends at the
video duration. Give the ends in increasing order.

Rules:

- A segment should normally cover at least 30 seconds.
- Split where the player's PURPOSE changes, not where the scenery changes.
- Name the intent that makes each whole segment coherent, in 6-20 words of concrete
  Minecraft vocabulary, grounded in the facts.
- Do not add narrative detail the evidence does not support.
- Cite the fact ids that support each segment.

VIDEO DURATION: <duration> s
FACTS (intent-neutral observations, video-relative seconds):
  [f12] 3.0-5.5s  <observation>
  ...
Return the JSON object with your segments.
\end{lstlisting}

\paragraph{Filtering.}
We clean the generated captions with simple rules:
\begin{itemize}[leftmargin=*,itemsep=1pt,topsep=2pt]
    \item decoding is constrained to a JSON schema;
    \item description times are clamped to their window;
    \item empty or zero-length items are dropped, 1.5\% of descriptions in the rewrite;
    \item task boundaries are closed into an exact partition on the frame clock.
\end{itemize}

\paragraph{Statistics.}
Table~\ref{tab:caption_stats} summarizes the corpus: 58{,}121 videos (2{,}943\,h), about 28M captions and 489M tokens. Descriptions and instructions overlap heavily by design. Their total span is about 6$\times$ the video duration, so every moment is described several times at several scales. Each task partition tiles its video exactly. 94.6\% of task intents and 84.9\% of instructions are unique strings.

\begin{table}[h]
\centering
\small
\caption{Caption corpus statistics for the VPT dataset. Spans are median [10th, 90th percentile]. Tokens are counted with the Cosmos3-Nano tokenizer.}
\label{tab:caption_stats}
\begin{tabular}{lrrcr}
\toprule
 & Count & Per video (median) & Span (s) & Tokens / caption \\
\midrule
Videos & 58{,}121 & -- & 241.0 (duration) & -- \\
Descriptions & 13{,}352{,}359 & 244 & 3.5 [1.0, 8.0] & 19.8 \\
Instructions & 13{,}155{,}310 & 237 & 3.5 [1.0, 8.0] & 15.2 \\
Task segments & 1{,}506{,}066 & 26 (5 partitions) & 31.5 [12.0, 62.6] & 16.1 \\
\midrule
Total tokens & \multicolumn{4}{r}{489.1M} \\
\bottomrule
\end{tabular}
\end{table}

\section{Implementation Details and Baseline Selection}
\label{app:baseline_details}

\subsection{Minecraft}

\subsubsection{V2A Implementation Details}
\label{app:v2a_impl}
\textbf{Architecture. } Unless noted otherwise, the V2A model for Minecraft uses an 80-million parameter V-JEPA~2.1 ViT-B model~\citep{assran2025v,MurLabadia2026VJEPA21} as a backbone. The model takes up to 8 history frames and 16 future frames as input, and predicts a chunk of up to 16 actions. The V2A predicts keyboard and mouse actions from the MineRL 1.0 action space~\citep{guss2019minerl}. The button actions are parameterized as categorical distributions, following the factorized button representation of VPT. The $x$ and $y$ offsets for mouse movement are parameterized as a 2D vector. For the diamond task, the final self-improving model is retrained with the larger V-JEPA~2.1 ViT-L encoder (Appendix~\ref{app:diamond_v2a}).

\textbf{Exploration Noise. } Exploration noise helps the agent learn novel interactions with the environment. However, if the noise is too high frequency, it may harm performance. Real-world actions are typically low frequency---a player will often repeat the same action, such as moving forward, over several time steps. We generate such low frequency exploration noise with random walks. We inject Brownian motion with standard deviation 1.0 for mouse exploration, and add noise corresponding to random walks of button presses for discrete actions.

\subsubsection{BASALT Evaluation}
Our method and the BASALT baselines were evaluated using the 11 world seeds from the BASALT 2022 challenge~\citep{Milani2023BASALTRetro}. Specific task prompts and environment settings can be found on the \href{https://minerl.readthedocs.io/en/latest/environments/basalt.html}{MineRL website}. The BASALT 2022 retrospective report~\citep{Milani2023BASALTRetro} contains comprehensive details and code describing the winning solutions to the challenge: GoUp, UniTeam, KAIROS, and others. 

While the BASALT criteria are intentionally fuzzy, we evaluate each method on objective criteria for fairness. For FindCave, success is granted when the agent enters a cave or ravine. On MakeWaterfall, we report success as when the player successfully places a water bucket, and ends the episode facing the waterfall to ``take a picture''. The waterfall must be in view during the episode's final frame for success to be granted. 
Because Mirror does not rely on action labels and its video planner makes the decisions, the planner also decides when an episode ends: it predicts a gray screen where human players pressed Esc. On MakeWaterfall, we therefore treat a gray screen predicted by the video planner, or reaching the time limit, as the agent pressing Esc. As the only exception, we ignore gray screens predicted before the agent has given one second of movement input (about four blocks of walking), to filter out false positives at the very start of an episode.

\subsubsection{Diamond Challenge}
\label{app:diamond}
\label{app:diamond_scope}
The following subsections describe the diamond experiments of Section~\ref{sec:oracle_survival}: the video planner's data (Appendix~\ref{app:diamond_vp_data}), the subgoal task planner (Appendix~\ref{app:task_planner}), the training of the self-improving V2A model (Appendix~\ref{app:diamond_v2a}), and the post-processing of the videos (Appendix~\ref{app:gui_postprocess}). Here we summarize what supervision enters the pipeline.

Neither the VP nor the self-improving V2A model is trained on expert actions, rewards, or privileged game state; the only actions the self-improving V2A model learns from are the ones the agent executed itself. The VPT action and event logs come in at two places. First, as the mouse cursor is missing inside menus in the VPT recordings, we draw it using the logged mouse position, as VPT's own data loader does~\citep{VPT2022OpenAI}. This only restores what the player saw on screen. Second, the logs help curate the VP's data. They direct the captioning VLM to salient moments (Appendix~\ref{app:vp_captions}) and mark where each subgoal of the diamond task happens (Appendix~\ref{app:diamond_vp_data}). This makes up for the limited capacity of our relatively small captioning model, the roughly 8B-parameter VLM of Cosmos3-Nano~\citep{Cosmos3}, which lacks Minecraft-specific domain knowledge. We believe that stronger VLMs with broader domain knowledge will make these log-based aids unnecessary. During rollouts, the subgoal task planner reads the agent's inventory, health, and whether a GUI is open. These state inputs are used only to adjust the VP's behavior by modulating its text prompt and guidance scale.

\subsubsection{Diamond Video Planner Data}
\label{app:diamond_vp_data}
Our VP was finetuned only on human gameplay from the VPT contractor recordings~\citep{VPT2022OpenAI}. Starting from the video planner pretrained on the VPT dataset (Appendix~\ref{app:VP}), we first finetune it for 30k steps on gameplay rendered with the post-processing of Appendix~\ref{app:gui_postprocess}, so that the planner gets used to the new cursor and hotbar. We then post-train it for about 50k steps on a curated VPT subset split into subgoals.

We split the task into 16 subgoals, from log to diamond (Figure~\ref{fig:task_planner}). In the 2{,}212 hours of VPT data versions 6.10 to 10, an inventory gain in the recorder log marks when a subgoal is completed, and a game statistic tells whether the item was mined, crafted, placed, or taken from a furnace. Each subgoal interval covers the activity leading up to the gain, such as the last run of attacks before a mined block or the time since the crafting screen opened. Training clips of 32\,s lie at least half inside an interval of their subgoal, and subgoals are sampled in proportion to the fourth root of their footage, which roughly balances them. The resulting clips span 665 hours, 403 of which fall inside subgoal intervals. Footage is plentiful for gathering (153 hours for cobblestone, 126 for log, and 68 for iron ore) but scarce for everything else: 9.5 hours for diamond and 2.7--12.3 hours for each crafting, placing, or smelting subgoal.

Each subgoal has a single caption of the form ``Current subgoal: \{item\}. \{instruction\}'', for example ``Current subgoal: stone pickaxe. Craft a stone pickaxe using cobblestone and sticks in the crafting grid.'' Captions carry no quantities, since the task planner decides when to move on. Prediction windows inside a subgoal interval get its caption, and some windows are left without text to train the unconditional branch for classifier-free guidance.

\subsubsection{Subgoal Task Planner}
\label{app:task_planner}
\begin{figure}[t]
\centering
\includegraphics[width=\linewidth]{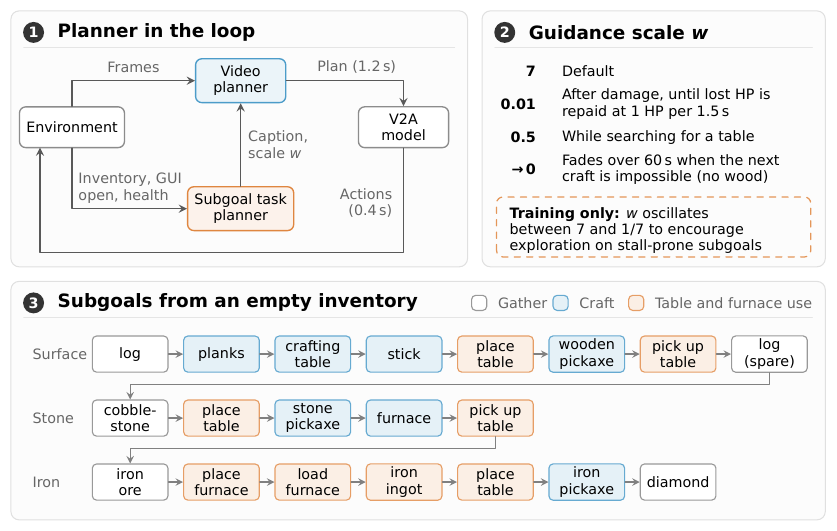}
\caption{\textbf{Subgoal task planner.} (1)~Once per plan, the planner reads the agent's inventory, health, and whether a GUI is open, and sends a subgoal caption and a guidance scale $w$ to the video planner; the V2A model never sees them. (2)~Guidance-scale rules; the dashed one is used only during training. (3)~Subgoals in the order the planner picks them from an empty inventory. Each 3$\times$3 craft happens at a crafting table placed for it and picked up afterwards.}
\label{fig:task_planner}
\end{figure}

Dreamer~V4~\citep{hafner2025dreamerv4} also conditions its agent on subgoals, prompting a one-hot task input through a fixed sequence of 25 tasks during evaluation. As the code is not public, we wrote our own heuristic subgoal planner (Figure~\ref{fig:task_planner}). 
Once per plan, it picks the first unmet subgoal by chaining backward from the diamond through the recipes, given the current inventory. It collects spare wood before going underground and enough cobblestone for both the stone pickaxe and the furnace, and it places a crafting table before every 3$\times$3 craft and picks it up afterwards, since it cannot tell whether an earlier table is still within reach. Its only inputs are the inventory counts, the health, and whether a GUI is open. Dreamer~V4 does not say how it decides that a subgoal is done, but its task set includes \texttt{open\_crafting\_table} and \texttt{open\_furnace}, and its evaluation prompts include \texttt{use\_stone\_pickaxe} and \texttt{use\_iron\_pickaxe} (Tables~4 and~6 of~\citet{hafner2025dreamerv4}). None of these can be read from the inventory or a GUI flag. Our planner uses no such game events and no learned classifier. The task planner's only outputs are the subgoal caption and the text CFG scale of the video planner. The V2A model never sees either. 

The text CFG scale is 7 by default. After the agent takes damage, it drops to 0.01 until the lost health is repaid at 1\,HP per 1.5\,s, so the video planner falls back on what players typically do when hurt, e.g., fight or flee.  In a fraction of the training episodes, it also oscillates between 7 and 1/7 on the subgoals where the agent stalls most (cobblestone, iron ore, loading the furnace, iron ingot, and diamond). Evaluation never uses this rule.

\subsubsection{Diamond V2A Training}
\label{app:diamond_v2a}
\paragraph{Online training.}
The self-improving V2A model starts from a pretrained V-JEPA~2.1 ViT-B encoder~\citep{assran2025v,MurLabadia2026VJEPA21} with freshly initialized action heads and learns online in 56 parallel environments. 
Training batches are sampled uniformly from a replay buffer that keeps the latest four in-game minutes of experience per environment. We keep the update-to-data ratio at about 1, lower than in typical sample-efficient RL methods that update many times per environment step. Our V2A objective has much lower variance than RL objectives, so the model learns efficiently from each transition, and compute is better spent on exploring new parts of the task than on replaying old experience. The V2A model's observations pass through a Wan2.2 VAE round trip, so that they look like the generated plans. Online training ran for 45k updates at batch 224, over 140 in-game hours.

\paragraph{Exploration.}
Exploration comes from two sources. The first is behavioral: lowering the text CFG scale pulls the plans toward the unconditional VP, which reproduces the exploratory and recovery behavior of human players. The second is action noise (Appendix~\ref{app:v2a_impl}), whose main role is to keep a minimum probability on every action. Actions with little or no visible effect, such as mouse clicks in an empty space, easily collapse to a single mode. The V2A model trains on its own executed actions and this results in a self-reinforcing collapse. The action exploration noise keeps a minimal floor probability, preventing any action from being starved.
Note that both kinds of exploration are double-edged: they diversify the data, but they also slow the agent down and lower its chance of reaching deep subgoals. We therefore explore mostly at the frontier, where the agent is believed to be stuck. In the last four minutes before each milestone deadline, the guidance scale oscillates between 7 and 0.05 on a log scale and the action noise turns on and off, each with a random period of 0.5 to 3 minutes. In a fraction of the training episodes, the task planner also oscillates the guidance scale on stall-prone subgoals (Figure~\ref{fig:task_planner}). Action noise is applied only during these exploration periods and the staggered entry below, and the V2A model trains on the noisy actions that were actually executed, rather than the clean actions it predicted. Even a small noise at this frontier is enough to prevent self-reinforcing collapse.
Finally, if all environments started their episodes at once, the replay would be dominated by whichever subgoal they were all working on. To balance the subgoals, we stagger the start of training: environment $g\in\{0,\dots,55\}$ first plays an unconditional world with action noise for $(g+\tfrac12)\cdot32$\,s (16\,s to 29.6\,min) before its first task episode, which spreads the environments across the task. This staggered entry is applied only once, when training starts.

\paragraph{Episodes and savepoints.}
To save compute, a training episode is split into segments that end at four milestones read from the inventory (a crafting table, a wooden pickaxe, a stone pickaxe, and three iron ore) and a final segment that ends at the diamond. The segments have limits of 10, 10, 10, 15, and 15 minutes plus the four-minute exploration period; a missed deadline or a third death ends the episode. We also save the world when an episode reaches the stone pickaxe, keep the 14 latest save points, and start a new episode from a random save with probability 0.25 times the filled fraction of this pool. These rules are not used at evaluation.
We note that these rules could tilt the training distribution through selection bias. The limits, however, are loose. During training, each limit was 1.9 to 2.8 times the mean time the agent took to reach the milestone, and at least 92\% of the reached milestones came within the limit. The deadlines thus rarely cut episodes that were still making progress and are unlikely to accelerate learning the way a reward would. If anything, the tilt works against speed, as an episode nearing its deadline enters the exploration period, which slows the agent down. Even if such a positive tilt exists, the four milestones are far coarser than the 15 to 20 rewarded subgoals of RL agents~\citep{VPT2022OpenAI,hafner2025dreamerv4}.
For the oracle VP, we do not use the milestone deadlines and keep only a global episode limit of up to 10 minutes. The oracle is an RL policy that progresses much faster than human players, so deadlines set for our VP, which follows the pace of human gameplay, are rarely reached.

\paragraph{Final model.}
After online training, we retrain a fresh V2A model with the larger V-JEPA~2.1 ViT-L encoder~\citep{assran2025v,MurLabadia2026VJEPA21} on the last 110 of the 140 in-game hours of online experience, excluding the first 30 hours, which are dominated by exploration. It trains on the agent's own executed actions for 20k steps at batch 128 and is the self-improving model in Figure~\ref{fig:minerl-progress}.

\subsubsection{GUI Video Post-Processing}
\label{app:gui_postprocess}
\begin{figure}[t]
\centering
\includegraphics[width=\linewidth]{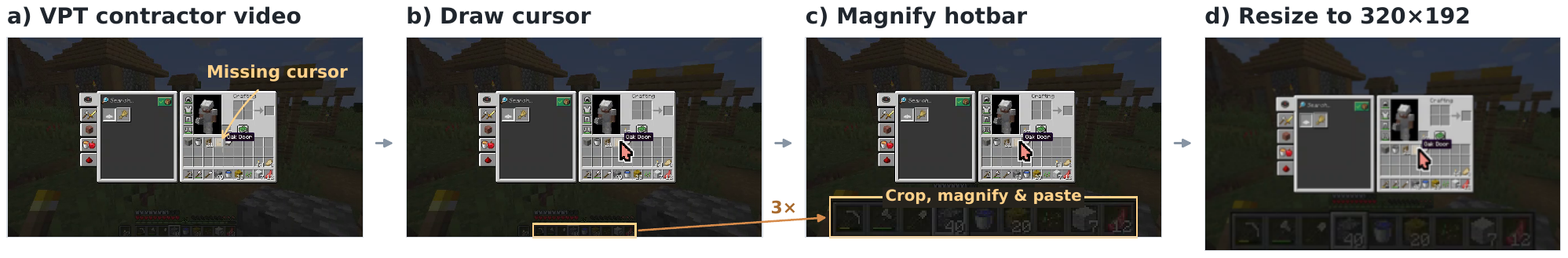}
\caption{\textbf{Video post-processing for the diamond task.} (a)~The original VPT video recordings do not include the cursor inside GUI menus. (b)~As in VPT's data loader~\citep{VPT2022OpenAI}, we draw the cursor from the mouse position log. However, VPT's original 16$\times$16 cursor is too small for the Wan2.2 VAE, which blurs or drops it (Figure~\ref{fig:vae_cursor}). Thus, for the diamond task we draw a high-visibility cursor: VPT's cursor enlarged 2$\times$, tinted coral, and outlined in black and white. (c)~The VAE also blurs the small item counts or turns them into other digits, so we crop the 184$\times$24-pixel hotbar from the video, enlarge it 3$\times$ with nearest-neighbor upsampling, and paste it back at the bottom center, moving the health and food bars up. (d)~Finally, we resize the $640\times360$ frames to $320\times192$ with bilinear interpolation and antialiasing. Of the filters we tried, it is the blurriest but keeps all task-relevant detail and causes the fewest VAE artifacts (reconstruction PSNR 32.7\,dB, versus 24.5--30.4\,dB for bilinear without antialiasing, bicubic, area, and Lanczos).}
\label{fig:gui_postprocess}
\end{figure}
\begin{figure}[t]
\centering
\includegraphics[width=\linewidth]{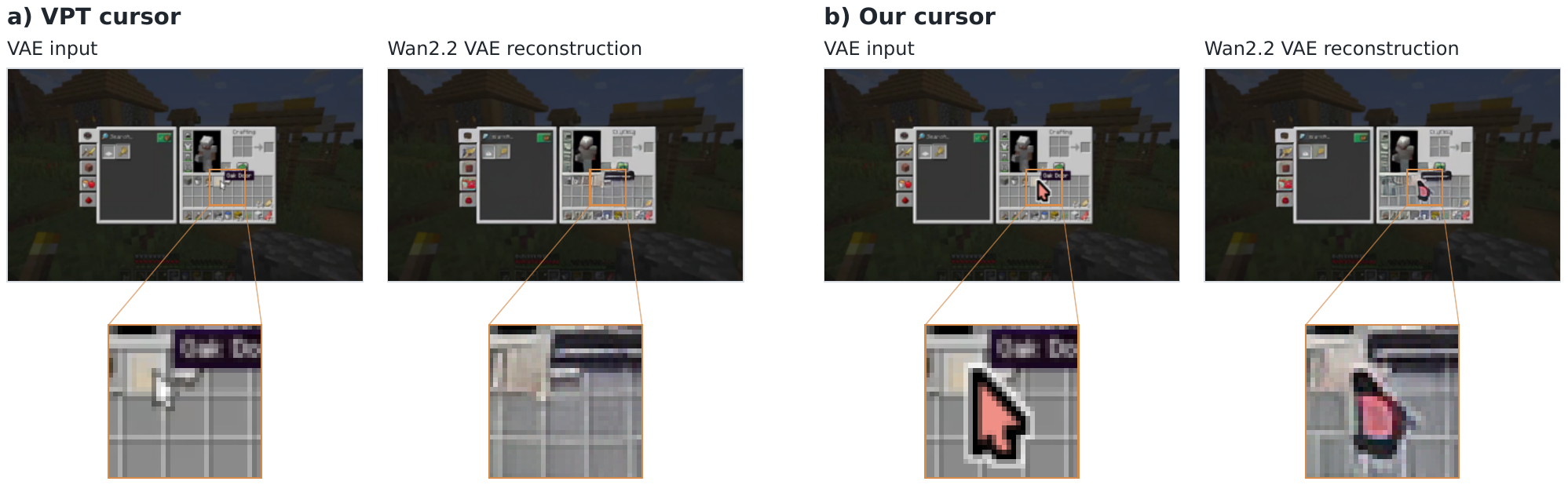}
\caption{\textbf{The Wan2.2 VAE loses the VPT cursor.} The same frame with (a)~VPT's cursor and (b)~the high-visibility cursor, before and after a round trip through the Wan2.2 VAE. VPT's small cursor nearly vanishes in the reconstruction, which is why we draw a larger, outlined cursor for GUI-heavy tasks.}
\label{fig:vae_cursor}
\end{figure}

These steps change only the video, not the game. As with the VPT recordings, the environment renders frames without a cursor and we draw ours on top, and we apply the same post-processing to the VP's training data and to the agent's observations during training and evaluation. The steps are needed because of the Wan2.2 VAE~\citep{wan2025wan}, whose latent space Cosmos3 is trained in. The VAE does not preserve small, high-precision details such as text and the cursor, which matter for computer-use tasks. With the oracle VP, which does not go through the VAE, the agent learns to mine diamonds without any of this post-processing. We therefore see the post-processing as a limitation of the Wan2.2 VAE rather than of our method, and we expect it to become unnecessary once video VAEs are trained on computer-use data.

\subsection{Robotics and Atari}

\subsubsection{Environments and Evaluation}

We evaluate on 13 DeepMind Control (DMC) tasks, four single-task OGBench environments, and seven Atari games. Each online method receives \(10^6\) environment actions and is evaluated every 25,000 actions over 20 episodes, with five training seeds per task. Random-V2A collects its interaction budget before offline fitting on different-sized subsets. We average tasks equally, reporting native return on DMC, success rate on OGBench, and \(100R_{\mathrm{agent}}/R_{\mathrm{expert}}\) on Atari.

\paragraph{OGBench.}
We use four visual single-task environments: \texttt{cube-single} and \texttt{cube-double} require rearranging one and two cubes; \texttt{scene} requires manipulating a drawer, window, and locking buttons; and \texttt{antmaze-large} requires navigating a quadruped robot to a goal. Observations are \(64\times64\) RGB images. We use task 2 for Cube Single, Cube Double, and Scene, and task 1 for AntMaze Large.

\paragraph{DeepMind Control.}
Following \citet{giammarino2024adversarialimitationlearningvisual}, we use 13 tasks from DMC~\citep{tassa2018deepmindcontrolsuite}: Ball-in-Cup Catch, Finger Spin, Cartpole Balance and Swingup, Pendulum Swingup, Walker Stand/Walk/Run, Hopper Stand/Hop, Quadruped Walk/Run, and Cheetah Run. Agents receive \(84\times84\) RGB observations with three-frame histories.

\paragraph{Atari.}
Following \citet{escontrela2023video}, we evaluate on Assault, Atlantis, Boxing, Freeway, Kangaroo, Pong, and Zaxxon from the Arcade Learning Environment~\citep{bellemare2013atari}. We use \(84\times84\) grayscale observations, four-frame histories, the minimal action set, sticky-action probability 0.25, and action repeat four. One million Atari decisions correspond to approximately four million emulator frames.

\subsubsection{Shared Implementation}

\paragraph{Action models.}
\methodname{}, BCO, Grounding, and Random-V2A use flow-based action models with method-specific conditioning. Continuous-control models use an IMPALA-style visual encoder and four 1024-unit flow layers. Atari uses a smaller encoder, one 512-unit flow layer, and factorized one-hot action codes projected to legal actions. Training uses Adam with learning rate \(3\times10^{-4}\), batch size 256, gradient clipping at 10, and random-shift augmentation. Action sampling uses ten Euler integration steps, from zero initialization at evaluation. LAIfO uses its separate single-action learner.

The action horizon \(K\) is five on OGBench; three on DMC except Cartpole Balance (seven), Cartpole Swingup (ten), and Pendulum Swingup (twenty); and two on Atari except Atlantis and Kangaroo (four).

\paragraph{Expert information.}
Experts are BRO~\citep{nauman2024biggerregularizedoptimisticscaling} on DMC, FQL~\citep{park2025flowqlearning} on OGBench, and DQN~\citep{mnih2013playingatarideepreinforcement} on Atari. BCO and LAIfO receive observation-only demonstrations: 50 episodes per Atari game and 1,000/500/500/50 episodes for Cube Single/Cube Double/Scene/AntMaze Large. On OGBench, LAIfO uses full-horizon trajectories that continue after success, whereas BCO uses the original shorter demonstrations. DMC uses 50 episodes for BCO and 100 for LAIfO. No method uses expert action labels, task rewards, or proprioceptive state.

\paragraph{Learned video planners.}
\methodname{} and Grounding use task-specific latent flow predictors with four residual blocks of width 512. A spatial autoencoder with encoder widths \(64,128,128,128\) maps frames at their native resolution to \(4\times4\times2\) latents on OGBench and DMC and \(7\times7\times2\) latents on Atari. The predictor models normalized latent differences from the current frame. Autoencoders train for 250,000 updates and predictors for \(10^7\) updates; flow training uses Adam with peak learning rate \(3\times10^{-4}\), cosine decay, batch size 256, gradient clipping at 1, and exponential-moving-average weights with decay 0.999, and we keep the checkpoints with the lowest validation loss. Prediction uses ten Euler steps and averages eight latent samples per predicted frame. Autoencoders are frozen during flow training, and both models remain frozen during downstream learning. Each task shares one planner checkpoint across downstream seeds. Atari planners use three context frames, while action models use four.

\subsubsection{Methods and Hyperparameter Selection}

\paragraph{\methodname{}.}
After 5,000 random actions, \methodname{} repeatedly obtains \(K\) future frames, predicts a \(K\)-action sequence from the current visual history and these frames, and executes it before replanning. Achieved observations and executed actions enter a 100,000-transition replay buffer, with 0.2 V2A model updates per environment action. Training actions are sampled stochastically from the flow policy on all benchmarks. On OGBench and DMC, each step also adds Gaussian noise to one randomly chosen action coordinate ($\sigma = 0.1$ on OGBench, task-specific on DMC). Atari relies on the stochastic flow sampling alone. Evaluation uses deterministic V2A model readout without collection noise.

For learned-planner experiments, we build the planner's training set by iterative dataset aggregation: starting from 200,000 expert transitions, each round executes the current planner--V2A controller and branches 20-step expert recoveries (30 steps in later rounds for three DMC tasks) from its states with probability 0.1 per step, with later rounds adding action noise and finally a mix of action corruptions (Gaussian noise, random, scaled, sticky, or delayed actions) to the executed controller actions. The final planner is trained from scratch on the aggregated corpus (3.5--13 million transitions per task, 5\% held out for validation). It conditions on three frames and predicts one future frame at a time, generating \(K\)-frame plans autoregressively.

\paragraph{Behavioral Cloning from Observation.}
BCO~\citep{torabi2018behavioralcloningobservation} alternates learner interaction, V2A model fitting, and policy fitting. The V2A model infers action sequences from expert observations, and an observation-conditioned policy learns these pseudo-labels and executes action chunks open loop. Both networks retain their parameters and optimizer state across rounds. Episode-level validation selects each fit, and pending fits finish before final evaluation.

We tune the initial collection size \(i_{\mathrm{pre}}\) and collection ratio \(\alpha\), giving approximately \(\alpha i_{\mathrm{pre}}\) actions between refits. Nine pairs are evaluated with three seeds on Cube Single, Finger Spin, and Pong, ranked by the final three evaluations. Selected \((i_{\mathrm{pre}},\alpha)\) values are \((40{,}000,0.1)\) for OGBench, \((20{,}000,0.2)\) for DMC, and \((20{,}000,0.4)\) for Atari.

\paragraph{LAIfO.}
LAIfO~\citep{giammarino2024adversarialimitationlearningvisual} learns an imitation reward by discriminating expert and learner observation transitions, then trains a single-action off-policy learner. Transition distance \(n\) controls discriminator inputs and critic bootstrapping. DMC uses released task-specific settings with our BRO experts; Atari uses a dueling Double-DQN \citep{wang2016duelingnetworkarchitecturesdeep, vanhasselt2015deepreinforcementlearningdouble} adaptation with greedy evaluation.

We tune exploration decay, transition distance, batch size, and a paired learning-rate/feature-dimension setting on Cube Single and Pong using 24- and 36-configuration grids, respectively, with three seeds per configuration. Selection uses performance at one million actions. Both suites select a 100k exploration decay, batch size 512, learning rate \(8\times10^{-5}\), and feature dimension 100. Transition distance is three for OGBench and one for Atari; discriminator learning rate is \(4\times10^{-4}\).

\paragraph{Grounding.}
Grounding~\citep{Luo2024GroundingVM} trains a goal-conditioned action model from random and plan-guided rollouts. Each training segment uses its final achieved image as the goal and executed actions as targets. Following an initial random-only fitting phase, replay sampling uses 30\% random and 70\% guided data. On OGBench, we follow the released LIBERO implementation, conditioning the model on a single current frame, with each guided collection following a single video plan without replanning. During evaluation, we periodically replan from the latest observation, with the number of plans set by each environment's time limit. For DMC and Atari, we use three and four frame observation histories, respectively. To accommodate their longer task horizon, we replan during both collection and evaluation after every two subgoals in oracle planner experiments and after every subgoal in generative planner experiments.

We tune random-action hold duration and optimizer-update intervals between guided and random collection using 27-configuration, three-seed sweeps on Cube Single, Finger Spin, and Pong, ranked by the final five evaluations. Selected \((\text{hold},\text{guided interval},\text{random interval})\) values are \((5,100,500)\), \((10,100,500)\), and \((5,200,1000)\) for OGBench, DMC, and Atari, respectively. OGBench collection intervals scale with task length.

Grounding's learned planners use the shared architecture and training recipe with task-specific subgoal targets. DMC and Atari reuse the aggregated expert/recovery corpus, conditioning on three frames to predict an expert-continuation subgoal \(K\) actions ahead. OGBench uses 1,000 expert episodes per task with a 950/50 training/validation split, conditions on one frame, and jointly predicts 7/14/20/70 subgoals for Cube Single/Cube Double/Scene/AntMaze Large. Following the original implementation, targets are spaced evenly from a sampled starting frame to shortly after expert success, with the final frame used for unsuccessful episodes.

\paragraph{Random-V2A.}
This baseline fits the same V2A model as \methodname{} independently on cumulative subsets of 100k to 1M transitions from a frozen uniformly random corpus. We hold out 5\% of episodes and select checkpoints by validation flow loss. Evaluation translates visual plans into action chunks using deterministic V2A readout; generative-VP results re-evaluate the same fitted models with the learned planner in place of the oracle.

\paragraph{VIPER.}
VIPER~\citep{escontrela2023video} learns a generative video model from expert observations and uses observation likelihoods as an RL reward. We include it as an Atari-only likelihood-reward reference, using estimates from its published results.

\section{Additional Results}
\label{app:additional_results}

\subsection{Minecraft}

\subsubsection{Data Scale}
We compare the data scale of our method with prior Minecraft agents in Table~\ref{tab:minerl_data_comparison}. The VPT BC and RL policies~\citep{VPT2022OpenAI} were trained on significantly greater amounts of data taken from internet videos. 

\subsubsection{BASALT Challenge}
Figure~\ref{fig:waterfall} contains an additional comparison to the BASALT challenge methods~\citep{Milani2023BASALTRetro}. Episodes from each BASALT baseline were taken directly from~\citet{milani2023bedd}. 

\begin{table}[t]
    \centering
    \caption{
        {Data scale of different Minecraft agents}
    }
    \label{tab:minerl_data_comparison}
    \scalebox{0.85}{
    \begin{tabular}{l c c c c c}
        \toprule
        \textbf{Method}
        & \textbf{Inputs}
        & \textbf{Outputs}
        & \multicolumn{3}{c}{\textbf{Minecraft Data (hours)}} \\
        \cmidrule(lr){4-6}
        & & & {Offline} & {Web} & {Online} \\
        \midrule

        Dreamer~V3
        & Video, inventory state
        & \begin{tabular}[c]{@{}l@{}}Keyboard, camera,\\ abstract crafting\end{tabular}
        & -- & -- & 1.4K \\

        Dreamer~V4
        & Video
        & Keyboard, mouse
        & 2.5K & -- & -- \\
        \midrule

        VPT (BC)
        & Video
        & Keyboard, mouse
        & 2K & 70K & -- \\
        
        VPT (RL)
        & Video
        & Keyboard, mouse
        & 2K & 70K & 230K \\
        \midrule

        Video Planner  (Ours)
        & Video, text prompt (optional)
        & Video
        & 3K & -- & -- \\
        
        V2A Model (Ours)
        & Video
        & Keyboard, mouse
        & -- & -- & 140 \\

        \bottomrule
    \end{tabular}
    }
\end{table}

\subsubsection{Diamond Challenge under the Dreamer~V3 Protocol}
\label{app:dreamerv3_protocol}
Figure~\ref{fig:minerl-progress} uses 60-minute episodes in which the player respawns after dying, as in VPT~\citep{VPT2022OpenAI} and Dreamer~V4~\citep{hafner2025dreamerv4}. Dreamer~V3~\citep{hafner2023dreamerv3}, and the PPO, Rainbow, and IMPALA results it reports, instead end each episode after 30 minutes or at the first death. Table~\ref{tab:dreamerv3_protocol} scores our self-improving agent under this protocol by cutting each of its 128 evaluation episodes at 30 minutes or at its first death, whichever comes first, which is equivalent to running them under Dreamer~V3's protocol. Among \methodname{}'s 128 episodes, 48 end early by death. \methodname{} still leads at every milestone except cobblestone. \methodname{} crafts an iron pickaxe in 8.6\% of episodes, against 1.7\% for Dreamer~V3, even though the baselines use task rewards, abstract crafting actions, and faster block breaking.

\begin{table}[h]
\centering
\caption{\textbf{Diamond Challenge under the Dreamer~V3 protocol.} Success rate (\%) per milestone after about 10M environment steps, with 30-minute episodes that end at the first death. Baseline values are read from Figure~9 of~\citet{hafner2023dreamerv3}; \methodname{} is evaluated on 128 episodes.}
\label{tab:dreamerv3_protocol}
\begin{tabular}{lrrrrr}
\toprule
Milestone & PPO & Rainbow & IMPALA & Dreamer~V3 & \methodname{} (self-improving) \\
\midrule
Log & 75.2 & 74.4 & 71.0 & 95.0 & \textbf{99.2} \\
Planks & 74.6 & 73.3 & 69.7 & 95.0 & \textbf{99.2} \\
Crafting table & 69.8 & 65.4 & 61.8 & 91.3 & \textbf{99.2} \\
Stick & 74.2 & 72.3 & 69.3 & 94.6 & \textbf{99.2} \\
Wooden pickaxe & 51.2 & 48.4 & 45.5 & 76.9 & \textbf{94.5} \\
Cobblestone & 26.2 & 23.9 & 31.0 & \textbf{67.4} & 57.0 \\
Stone pickaxe & 5.7 & 4.7 & 10.1 & 39.5 & \textbf{53.1} \\
Iron ore & 0.4 & 0.7 & 1.4 & 12.7 & \textbf{22.7} \\
Furnace & 14.1 & 11.0 & 18.9 & 51.7 & \textbf{53.1} \\
Iron ingot & 0.0 & 0.3 & 0.9 & 7.0 & \textbf{18.8} \\
Iron pickaxe & 0.0 & 0.0 & 0.0 & 1.7 & \textbf{8.6} \\
Diamond & 0.0 & 0.0 & 0.0 & 0.0 & 0.0 \\
\bottomrule
\end{tabular}
\end{table}

\subsubsection{Text-Conditioned Minecraft Evaluation}
\label{app:minecraft-text}
\begin{figure}[h]
    \includegraphics[width=\linewidth]{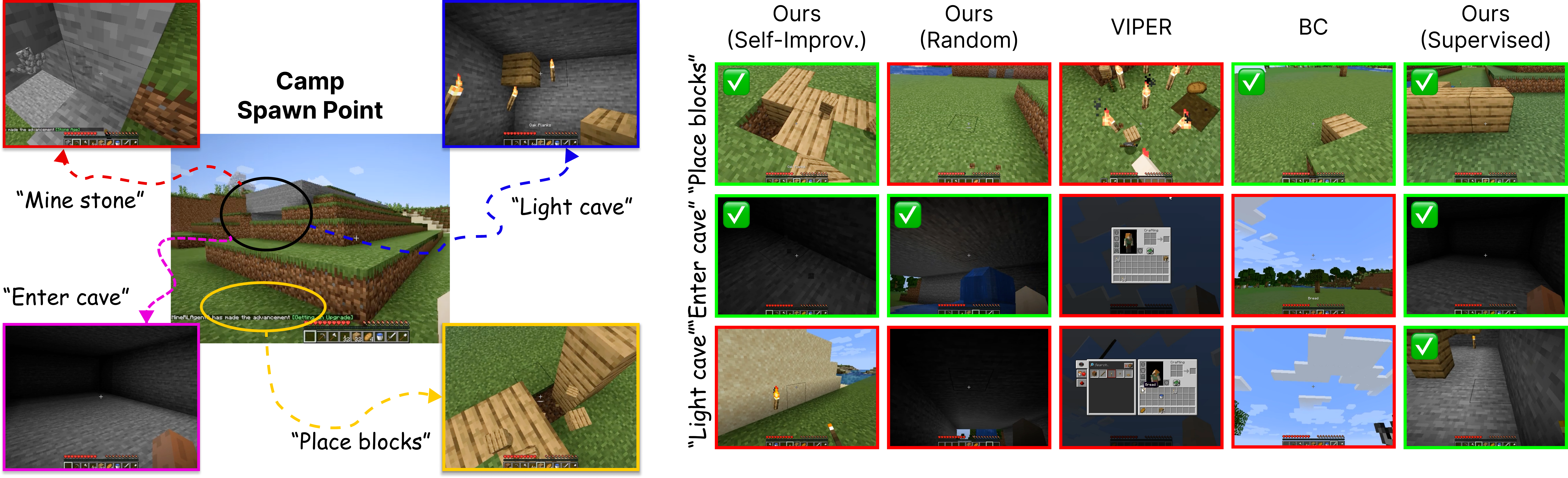}
    \caption{\textbf{Text-conditioned planning on Camp.} We demonstrate text-conditioned planning on the camp environment. While our self-improving approach is able to solve many tasks, it struggles on tasks with fine-grained environment interactions such as placing torches in the cave. Meanwhile, VIPER, our action-free baseline, struggles to learn the action space of Minecraft. BC solves some tasks, but remains limited in its generalization ability.  }
        \label{fig:minecraft-text-camp}

\end{figure}

We evaluate text control in two worlds: Woodland and Camp. The worlds are designed to contain various locations of interest, such as trees, a pool, a hill, and a cave. In each world, the player starts with an iron sword, iron tools, torches, wood planks, bread, and a water bucket. We evaluate 10 episodes per text prompt. In each episode, we first make the player pan the camera left and right to observe their environment. Then it must complete each task given by the text prompt within 30 seconds. The video planner uses a classifier-free guidance scale of 7.0. We include more results on text-conditioned task guidance in Figure~\ref{fig:minecraft-text-camp}.

\subsection{OGBench Generalization Analysis}
\label{app:ogbench_generalization_analysis}
\begin{table}[t]
\centering
\caption{\textbf{Transfer to unseen OGBench goals.} Success rate (\%, mean $\pm$ SE across five seeds), averaged over target goals. All methods use target-task oracle planners; BCO-V2A uses BCO's saved V2A model.}
\label{tab:goal_transfer}
\vspace{6pt}
\small
\begin{tabular}{lcccc}
\toprule
Environment & Ours & Random-V2A & BCO-V2A & Grounding \\
\midrule
Cube Single   & \textbf{78.3$\pm$2.8} & 25.3$\pm$4.2 & 8.5$\pm$2.5 & 5.0$\pm$0.8 \\
Cube Double   & \textbf{27.0$\pm$4.6} & 20.7$\pm$2.8 & 2.0$\pm$1.6 & 1.0$\pm$0.7 \\
Scene         & 32.5$\pm$3.5 & \textbf{50.0$\pm$5.4} & 1.5$\pm$1.5 & 3.0$\pm$0.5 \\
AntMaze Large & \textbf{25.8$\pm$1.5} & 0.0$\pm$0.0 & 0.0$\pm$0.0 & 0.0$\pm$0.0 \\
\bottomrule
\end{tabular}
\end{table}

Table~\ref{tab:goal_transfer} evaluates transfer to unseen OGBench goals, with all methods using oracle planners for the target tasks. Our V2A model outperforms Random-V2A in three of four environments and BCO-V2A, the V2A model learned by BCO, in all four. These results suggest that video-guided exploration helps the model acquire behaviors that remain useful beyond the tasks used for data collection. Grounding achieves at most 3.8\% success, suggesting that learning to translate frame-level transitions into actions supports stronger task generalization than learning a goal-conditioned policy. 

\subsection{Individual Task Learning Curves for Robotics and Atari}
\label{app:robotics_atari_individual_task_curves}

\begin{figure*}[t]
  \centering
  \includegraphics[width=0.9\textwidth]{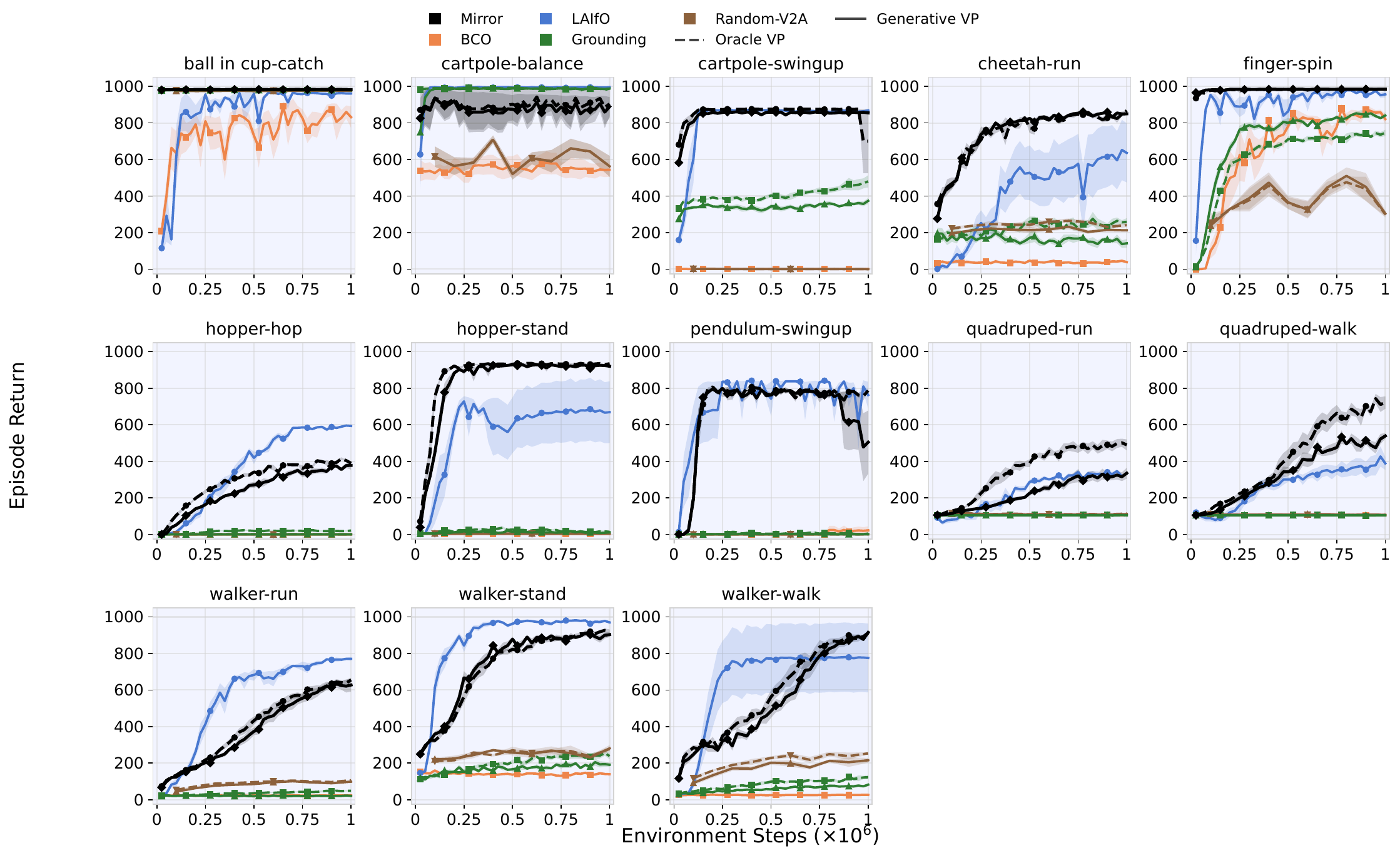}
  \caption{\textbf{Per-task performance on DeepMind Control.}
  Episode return on 13 tasks. Curves show the mean $\pm$ one standard
  error over three seeds.}
  \label{fig:dmc_individual}
  \vspace{-1\baselineskip}
\end{figure*}

\begin{figure*}[t]
  \centering
  \vspace{-1\baselineskip}
  \includegraphics[width=0.9\textwidth]{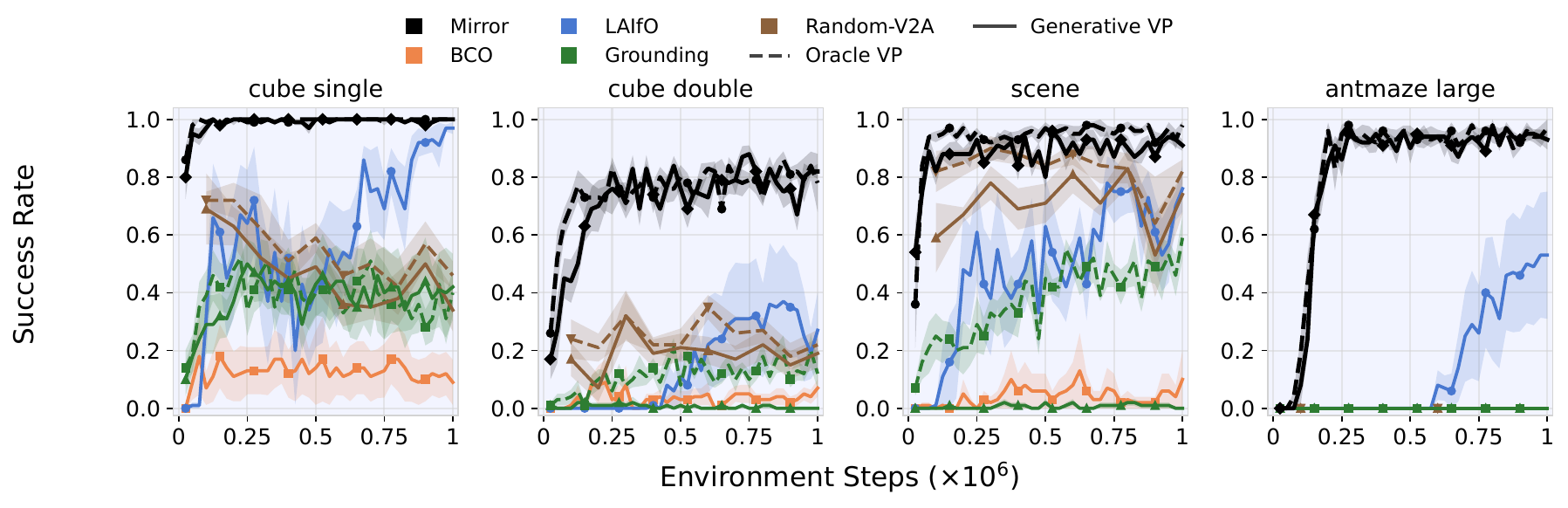}
  \caption{\textbf{Per-task performance on OGBench.}
  Success rate on four single-task environments. Curves show the mean
  $\pm$ one standard error over three seeds.}
  \label{fig:ogbench_individual}
  \vspace{-1\baselineskip}
\end{figure*}

\begin{figure*}[t]
  \centering
  \vspace{-1\baselineskip}
  \includegraphics[width=0.9\textwidth]{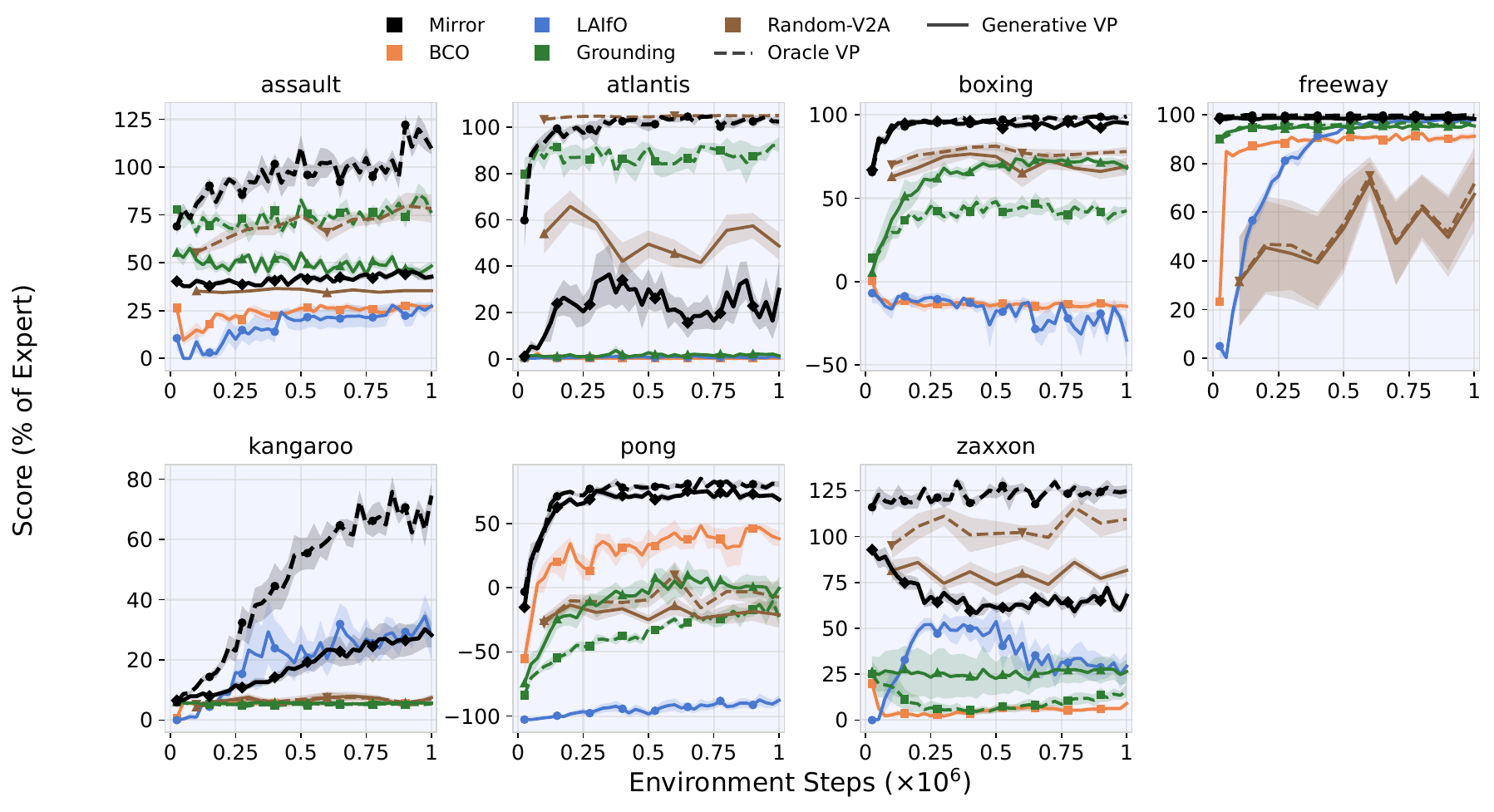}
  \caption{\textbf{Per-game performance on Atari.}
  Scores on seven games, expressed as a percentage of each game's expert
  score. Curves show the mean $\pm$ one standard error over three seeds.}
  \label{fig:atari_individual}
  \vspace{-1\baselineskip}
\end{figure*}

\begin{figure}[t]
    \centering
    \includegraphics[width=\linewidth]{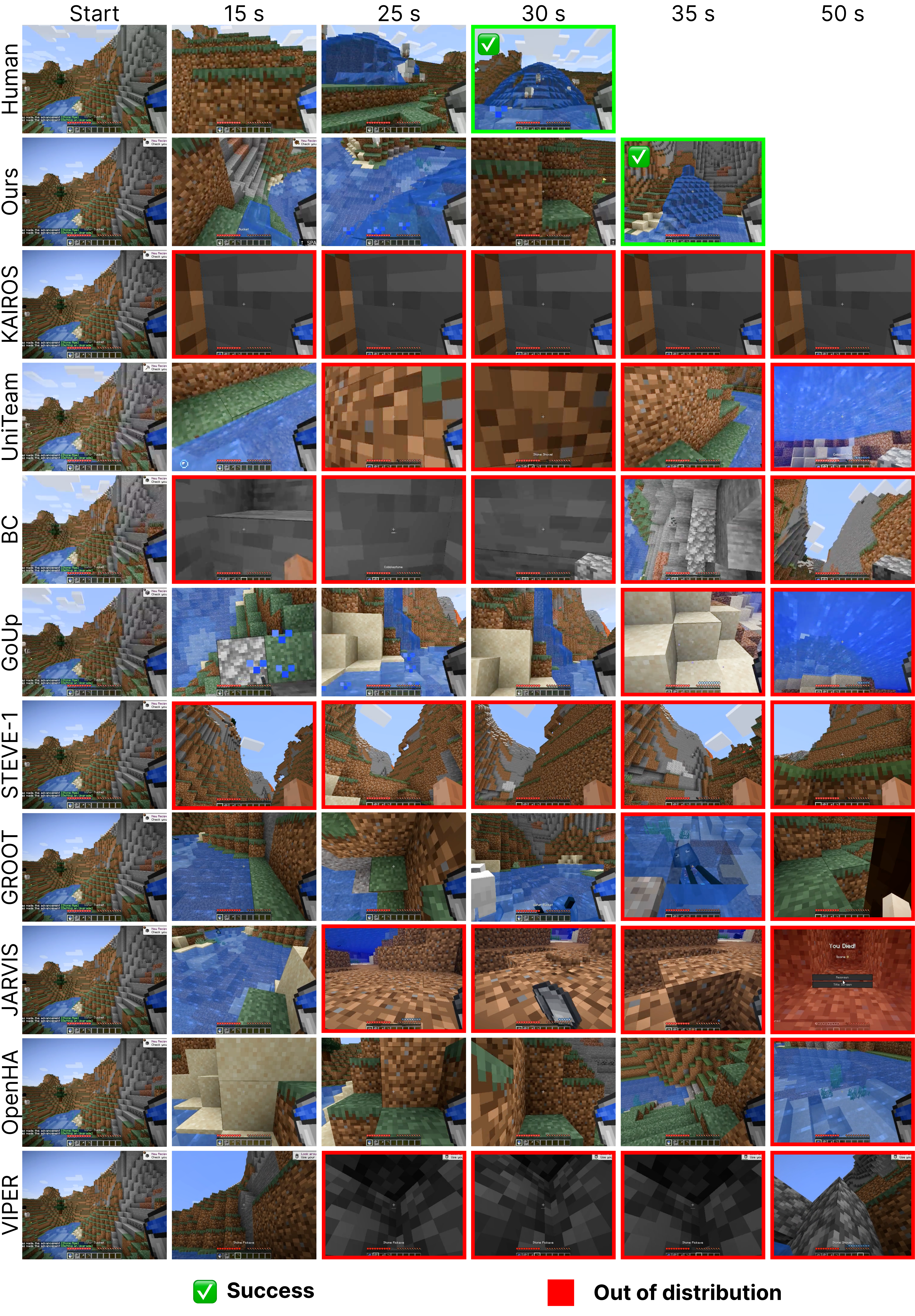}
    \vspace{-2\baselineskip}
    \caption{\textbf{MakeWaterfall comparisons.} \methodname{} is capable of the long horizon planning necessary for MakeWaterfall. The agent must make a waterfall, remember its location, and move to another location to take a picture of it. Even though BC, UniTeam and KAIROS are trained with action labels, they achieve 0\% success on this long-horizon task. They often go out of distribution. 
    GoUp, which relies on scripted actions, cannot adapt to unseen scenarios such as when underwater. While it is able to place the water bucket with a pre-programmed instruction, it does not know what to do when underwater. The VLA approaches, STEVE-1, GROOT, JARVIS-VLA, and OpenHA, struggle at following the task instructions. VIPER, our action-free baseline, struggles to learn the Minecraft action space.}
    \label{fig:waterfall}
\end{figure}

We report per-task results for DeepMind Control, OGBench, and Atari in Figures~\ref{fig:dmc_individual}, \ref{fig:ogbench_individual}, and \ref{fig:atari_individual}, respectively.

\newpage
\vspace{-5\baselineskip}

\end{document}